\documentclass[12pt]{article}
\usepackage[table,xcdraw]{xcolor}
\usepackage{setspace}  
\usepackage{apacite}
\usepackage{natbib}
\usepackage{graphicx}
\usepackage{booktabs}
\usepackage{pdfpages}
\usepackage{textcomp}
\usepackage{geometry}
\usepackage{caption}
\usepackage{amsmath}
\usepackage{etoolbox}
\usepackage[normalem]{ulem}
\useunder{\uline}{\ul}{}
\usepackage{longtable}
\usepackage{tabu}
\usepackage{multirow}
 
 \AtBeginEnvironment{longtable}{\singlespacing}
\newcommand\token{\textsc{token}}
\newcommand\tokens{\textsc{tokens}}
\newcommand\lemma{\textsc{lemma}}
\newcommand\lemmas{\textsc{lemmas}}

\makeatletter
\def\new@fontshape{}
\makeatother
\usepackage{gb4e}
\noautomath

\makeatletter
\renewcommand{\thexnumi}{\if@noftnote\@xsi{xnumi}\else\roman{xnumi}\fi}
\makeatother
\begin{document}

\title{The Role of Verb Semantics in Hungarian\\Verb-Object Order} 
\date{June 11, 2020}
\author{Dorottya Demszky\thanks{Contact: ddemszky@stanford.edu}, L\'aszl\'o K\'alm\'an, Dan Jurafsky, Beth Levin}
\maketitle
\begin{abstract}
    Hungarian is often referred to as a discourse-configurational language, since the structural position of constituents is determined by their logical function (topic or comment) rather than their grammatical function (e.g., subject or object). We build on work by Koml\'osy (1989) and argue that in addition to discourse context, the lexical semantics of the verb also plays a significant role in determining Hungarian word order. In order to investigate the role of lexical semantics in determining Hungarian word order, we conduct a large-scale, data-driven analysis on the ordering of 380 transitive verbs and their objects, as observed in hundreds of thousands of examples extracted from the Hungarian Gigaword Corpus. We test the effect of lexical semantics on the ordering of verbs and their objects by grouping verbs into 11 semantic classes. In addition to the semantic class of the verb, we also include two control features related to information structure, object definiteness and object NP weight, chosen to allow a comparison of their effect size to that of verb semantics. Our results suggest that all three features have a significant effect on verb-object ordering in Hungarian and among these features, the semantic class of the verb has the largest effect. Specifically, we find that stative verbs, such as \emph{fed} `cover', \emph{jelent} `mean' and \emph{\"ovez} `surround', tend to be OV-preferring (with the exception of psych verbs which are strongly VO-preferring) and non-stative verbs, such as \emph{b\'ir\'al} `judge', \emph{cs\"okkent} `reduce' and \emph{cs\'okol} `kiss', verbs tend to be VO-preferring. These findings support our hypothesis that lexical semantic factors influence word order in Hungarian.

\end{abstract}
\doublespacing
\section{Introduction}
\label{sec:intro}

Hungarian is often referred to as a free word-order language, given that grammatical functions, such as subject and object, are not linked to invariant structural positions in the sentence \citep[p.~2]{kiss2002syntax}. For example, all six permutations of the transitive verb \emph{szeret} `love', subject \emph{J\'{o}zsi} `Joe' and object \emph{S\'{a}rit} `Sarah-\textsc{acc}' are possible in principle, as shown in (\ref{ex:cat}).

\begin{exe}
    \ex \begin{xlist}
        \ex {\gll J\'{o}zsi szereti S\'{a}rit. \\
        Joe loves Sarah-\textsc{acc}\\
        \glt `Joe loves Sarah.'}
        \ex J\'ozsi S\'arit szereti.
        \ex S\'arit szereti J\'ozsi.
        \ex S\'arit J\'ozsi szereti.
        \ex Szereti S\'arit J\'ozsi.
        \ex Szereti J\'ozsi S\'arit.
    \end{xlist}
\label{ex:cat}
\end{exe}

Despite the apparent freedom of word order, it is widely accepted that information structure plays a major role in determining the word order of a Hungarian sentence \citep[see][among others]{kiss1978magyar,kiss1981structural,kiss2002syntax,kalman1989magyar,gecseg2009new}. For example, only orders (1b) and (1c), where `Sarah' is in the preverbal focus position, can be used in felicitous replies to the question  `Who does Joe love?'. Examples such as these suggest that the discourse context is crucial for determining the felicity of particular word orders.

In this paper we argue that, in addition to information structure, the lexical semantics of the verb also plays a significant role in determining Hungarian word order.  Our hypothesis builds on the work of \citet{komlosy1989fokuszban}, who presents evidence that certain verbs and their objects show systematic ordering preferences in Hungarian across discourse contexts. His findings, involving approximately 80 verbs, suggest that lexical semantic factors such as verb choice might also be at play in determining Hungarian word order. While the nature of this lexical influence has not yet been investigated, there are semantic similarities among verbs that according to Koml\'osy's study share the same ordering preference.  For example, many verbs he identifies as having a preference to follow their objects (OV-preferring) are psych verbs (e.g., \emph{ut\'{a}l} `hate', \emph{tud} `know', \emph{eml\'ekszik} `remember') and many verbs he identifies as OV-preferring express states or spatial configuration (e.g., \emph{tartalmaz} `contain', \emph{marad} `remain'). Given this, we hypothesize that the systematic ordering preference of verbs is driven by lexical semantic factors.

In order to investigate the role of lexical semantics in determining Hungarian word order, we conduct a large-scale, data-driven analysis on the ordering of 380 transitive verbs and their objects, as observed in hundreds of thousands of examples extracted from the Hungarian Gigaword Corpus \citep{oravecz2014hungarian}. We use empirical methods, as they allow us to verify and estimate the significance of word order tendencies using a large corpus of natural language. Such methods have not been used previously to study Hungarian word order, so we draw on works in other languages \citep{bresnan2007predicting,benor2006chicken}, which seek to estimate the influence of different factors on word order patterns via a logistic regression model trained on a large corpus. We restrict our analysis to transitive verbs and their objects to reduce the number of confounds arising from verbs with multiple arguments in the verb phrase.

We test the effect of lexical semantics on the ordering of verbs and their objects by grouping verbs into semantic classes. We identify 11 semantic classes based on salient semantic patterns in our training data as well as in previous literature on verb classes \citep[][]{levin1993english}, and we use these classes as features in our analysis. In addition to the semantic class of the verb, we also include two control features related to information structure, chosen to allow a comparison of their effect size to that of verb semantics. These features are object definiteness and object NP weight. \citet{komlosy1989fokuszban} has noted the influence of definiteness on verb-argument ordering, while the influence of constituent weight on word order has been noted for other languages \citep{wasow1997remarks}.  We obtain object definiteness and object NP weight automatically, using natural language processing tools for Hungarian \citep{qi2018universal,tron2005hunmorph}. We construct a logistic regression model to study the effect of these three features (semantic class of the verb, object definiteness and object NP weight) on the ordering of verbs and objects in our data.




Our results suggest that the features we used have a significant effect on verb-object ordering in Hungarian and among these features, the semantic class of the verb has the largest effect. Specifically, we find that stative verbs tend to be OV-preferring (with the exception of psych verbs which are strongly VO-preferring) and non-stative verbs verbs tend to be VO-preferring. These findings support our hypothesis that lexical semantic factors influence word order in Hungarian.

The organization of the paper is as follows. In Section~\ref{sec:research_questions}, we lay out our research questions, which we group into three topically related sets. In Section~\ref{sec:background}, we present background literature on Hungarian.  We then describe our methods for answering the research questions in Section~\ref{sec:methods}: the data and the pre-processing (Section~\ref{ssec:data}), the verb classification (Section~\ref{ssec:verb_classes}) and other, object-related features we used (Section~\ref{ssec:object_features}) and finally, the logistic regression model (Section~\ref{ssec:model}). We return to the research questions again in Section~\ref{sec:results}, where we present and discuss our results. In our conclusion (Section~\ref{sec:conclusion}), we summarize our work and discuss avenues for future research.

\clearpage
\section{Research Questions}
\label{sec:research_questions}

In Section~\ref{sec:intro}, we introduced our main motivation for focusing on the role of three features on Hungarian verb-object order. Here we provide our three sets of topically-related research questions and a glance at the methods we use to address them. 

\begin{enumerate}
    \item \emph{Ordering Preference of Verbs}\\How well can we predict the ordering of verbs and their objects in our data based on the verb exclusively?  Which transitive verbs have an OV preference, which ones have a VO preference, and which show no preference? 
    
    \underline{Methods:} We extract verb-object pairs from our corpus and run a logistic regression model to predict their ordering based on the lemma of the verb exclusively.

    \item \emph{Verb Classes}\\Can we identify a small set of semantic classes for the verbs, which are salient based on previous literature and based on the semantic features that seem relevant for the verbs' ordering preference? How well can we predict the ordering of verbs and objects based on the semantic class that the verb belongs to? Which verb classes have a VO preference, which ones have an OV preference and which none?
    
    \underline{Methods:} We identify a set of semantic classes and assign verbs to them. We then run a logistic regression model to predict the ordering of verbs and their objects based on the verb's semantic class. During prediction, we test on previously unseen data that we did not look at while identifying the verb classes.

    \item \emph{Importance of Object-Related Features}\\Besides the verb's semantic class, how well can object-related features, and specifically the definiteness of the object and the weight of the object NP, predict the relative ordering of the verb and the object?
    
    \underline{Methods:} Extract the object-related features automatically from the corpus, for each verb-object pair. Run a logistic regression model to predict the ordering of verbs and objects based on each of these features separately. 
    
\end{enumerate}
\section{Background on Hungarian}
\label{sec:background}

In this section, we discuss a few aspects of Hungarian that have been investigated in previous work and relate to word order. These aspects include the discourse-configurationality of Hungarian (Section~\ref{ssec:discourse_configurationality}), object definiteness (Section~\ref{ssec:obj_definiteness}), noun incorporation (Section~\ref{ssec:noun_incorporation}), the information content of verbs (Section~\ref{ssec:info_verbs}) and prosodic classification of verbs (Section~\ref{ssec:verb_prosody}).

\subsection{Discourse-configurationality}
\label{ssec:discourse_configurationality}

As we mentioned, it is widely accepted that in Hungarian, the syntactic structure of a sentence is determined by its information structure --- in other words, Hungarian is considered by many to be a discourse-configurational language \citep{kiss1995discourse}, even though there are disagreements as to what this term entails \citep{gecseg2009new}. Reviewing the extensive literature on Hungarian syntax is beyond the scope of this paper. Here we provide a brief overview of some of the key properties of Hungarian syntax that are relevant to the study of verb-object ordering. 

At a high level, a Hungarian sentence is divided into a logical subject (which some call a topic) and a logical predicate (which some call a comment), in this order. The preverbal part of the predicate, following the logical subject, is also called the ``focus field'' \citep{brody1990some} and it is reserved for the focus constituent. A variety of different elements can occupy the focus position, including preverbal particles, negative quantifiers, bare NPs and focused definites and indefinites. Generally, these elements are considered to be in complementary distribution \citep{gecseg2009new}, but there is evidence suggesting that their patterning is more complex \citep{farkas1986syntactic,kiss2002syntax}. Related to the positions of different grammatical elements in the sentence, there is also disagreement as to whether Hungarian has a neutral word order (SVO) \citep{kiefer1967emphasis,kalman1986hocus,maracz1989asymmetries} or whether there is no neutral word order in Hungarian \citep{kiss1981structural,kiss2002syntax}. Given the close and complex interaction between semantics, pragmatics and word order in Hungarian, many aspects of the theory of Hungarian syntax (e.g., the definition of a logical subject or what it means for a sentence to be have ``neutral'' word order) are much contested.

In this paper, we study the relative position of the verb and its object, given that the work of \citet{komlosy1989fokuszban} suggests that the verb plays a role in the word order of the sentence. The most relevant aspect of Hungarian syntactic theory to us is that when the object is preverbal, it can either be in the topic or focus position. The examples\footnote{The examples are based on the ones presented in different chapters of \citet{kiss2002syntax}.} in (\ref{ex:topic_focus}) show the possible positions that the grammatical object \emph{S\'arit} `Sarah-\textsc{acc}' can occupy. The object is preverbal both when it is the topic of the sentence (\ref{ex:topic_focus}a) and when it is focused (\ref{ex:topic_focus}b) and it is postverbal when it is neither focused nor topicalized (\ref{ex:topic_focus}c). 

\begin{exe}
\ex \begin{xlist}
    \ex {\gll \lb{Topic}S\'arit] \lb{Pred}el\"ut\"otte egy aut\'o] \\
    Sarah-\textsc{acc} \quad\quad\quad hit a car \\
    \glt `A car hit Sarah. [Sarah was hit by a car]'}
    \ex {\gll \lb{Topic}J\'ozsi] \lb{Pred}\lb{Focus}S\'arit] szereti] \\
     \quad\quad\quad Joe \quad\quad\quad\quad\quad Sarah-\textsc{acc} loves\\
    \glt `It is Sarah whom Joe loves.'}
    \ex {\gll \lb{Topic}J\'ozsi] \lb{Pred}szereti S\'{a}rit] \\
     \quad\quad\quad Joe \quad\quad\quad loves Sarah-\textsc{acc}\\
    \glt `Joe loves Sarah.'}
\end{xlist}
\label{ex:topic_focus}
\end{exe}

In our analyses, we do not distinguish between topicalized and focused objects, nor between their contrastive vs. non-contrastive subcategories (another distinction that is beyond the scope of this work). Identifying the discourse function would require manual labeling, since there is no reliable automatic method for doing so. Moreover, these discourse functions can be challenging to even label manually based on text, since intonation plays a key role in their interpretation \citep{kiss2002syntax}. Therefore, we do not incorporate discourse function into our analyses, which is a limitation of this work. However, we believe that studying the ordering preference of verbs, even without knowing the information structure of the sentence, is valuable as we can learn whether and to what extent verb meaning has an influence on word order that persists across discourse contexts.\footnote{We believe that incorporating discourse context would only strengthen our findings since doing so would allow us to remove examples where the object is in the contrastive topic or the contrastive focus position, given that contrastive constructions are considered less ``neutral'' due to the added emphasis on the contrastive element.}

\subsection{Object Definiteness}
\label{ssec:obj_definiteness}
In Hungarian, the definiteness of an object is closely linked to its discourse function, with definite objects being more likely to represent known information and indefinite objects being more likely to represent new information. Therefore, the definiteness of the object can be predictive of its syntactic position. Specifically, since the focus position is reserved for non-presupposed information \citep{kiss2002syntax}, indefinite objects are more likely to occupy this slot than definite ones. In contrast, definite objects are more likely to follow the verb, or occasionally to be topicalized. Even though the close relationship between the definiteness and the syntactic position of objects is assumed and/or mentioned in the literature on Hungarian syntax, the extent to which object definiteness is predictive of verb-object ordering has not been studied. We incorporate object definiteness as a feature into our predictive model to compare its effect to that of verb semantics.

Hungarian is well known for marking the definiteness of the object on the verb. Specifically, Hungarian has two inflectional paradigms, the objective and subjective conjugations. The objective conjugation signals the presence of a definite object. These conjugations are relevant for us because they allows us to detect the definiteness of the object based on the morphological features of the verb.

\subsection{Noun Incorporation}
\label{ssec:noun_incorporation}
Bare objects, which constitute a special subcategory of indefinite objects, are in fact required to be in the preverbal focus position unless they is another element in contrastive focus. Such preverbal bare NPs are considered to be instances of noun incorporation and their syntactic behavior is very similar to that of preverbs, forming a semantic and intonational unit with the verb \citep{kiefer1990noun,farkas2003semantics}. Since the relative ordering of bare nouns and verbs is not flexible, unlike the relative ordering of non-bare objects, we remove examples with bare objects from our analyses.

\subsection{The Information Content of Verbs}
\label{ssec:info_verbs}

Given the close relationship between information structure and word order in Hungarian, it is natural to wonder whether the information content of the verb (in the particular context) plays an important role in Hungarian word order. For example, do verbs with relatively low information content (so called ``light'' verbs), such as \emph{tesz} `make/put' and \emph{ad} `give', tend to follow their objects and verbs with relatively high information content tend to precede their objects? To our knowledge, this question has not been empirically investigated, and we also leave this question for future work to address, given the reasons outlined below.

First, the information content of the verb is heavily context dependent. Factors that may influence the informativeness of the verb include the discourse context and the nature of the arguments (e.g., their grammatical function and semantic properties). The influence of the arguments is why previous work studies the lightness of verbs in particular \emph{constructions} rather than the lightness of verbs in isolation \citep{vincze2010hungarian}.\footnote{\citet{vincze2010hungarian} define light verb construction based on \citet{sag2002multiword} as a type of lexicalized phrase or flexible expression that is neither idiomatic nor productive, and its meaning is not completely compositional.  Noun + verb combinations, such as \emph{bejelent\'est tesz} `make an announcement', are one category of light verb constructions they recognize.} As we discussed in Section~\ref{sec:intro}, we are interested in the influence of verb meaning on verb-object ordering across contexts, and we do not consider context-dependent features in this study.

Second, due to our experimental design, our dataset only has a limited number of verbs that occur in light verb constructions identified by \citet{vincze2010hungarian}. This is because canonical examples of light verb constructions, as reported by \citet{vincze2010hungarian}, involve bare nouns and verbs.  We remove such constructions from our analyses for the reasons described in Section~\ref{ssec:noun_incorporation}. Moreover, several light verbs that occur in light verb constructions tend to take multiple arguments (e.g., \emph{tesz} `make/put' frequently occurs with directional arguments). We remove verbs that frequently take multiple arguments from our data to control for confounds (e.g., the influence of the other argument on the ordering of the object and the verb). Due to the limited presence of light verbs, our data is not best suited for studying the influence of verb information content on verb-object ordering.

Even though studying the influence of verb informativeness on verb-object ordering is beyond the scope of this current work, we believe that it is a potentially relevant factor and we plan to address this question in future work.






\subsection{Verbs Sharing an Ordering Preference}
\label{ssec:verb_prosody}
As discussed in Section~\ref{sec:intro}, our work builds on that of \citet{komlosy1989fokuszban}, who identifies systematic ordering preferences for a group of Hungarian verbs. Specifically, Koml\'osy describes the verbs' prosodic behavior, which maps onto their ordering preferences, given that verbs that are on the left edge of the predicate (preceding their objects) tend to carry the main sentential stress.\footnote{In Section~\ref{ssec:discourse_configurationality} we mentioned that there is a close relationship between the intonation of a sentence and its information structure. The main sentential stress in a Hungarian sentence falls on the first major constituent in the predicate. Therefore, if there is a preverbal argument, it carries the main sentential stress as opposed to the verb.} He found that certain verbs tend to avoid carrying the main sentential stress (e.g., \emph{tal\'al} `find', \emph{tartalmaz} `contain', \emph{marad} `remain') while others tend to seek it (e.g., \emph{ut\'{a}l} `hate', \emph{tud} `know', \emph{eml\'ekszik} `remember'). He classifies such verbs into two classes, which he calls stress-avoiding and stress-seeking verbs, respectively, and classifies verbs with no strong preference as regular verbs. \cite{kalman1986hocus} further break down Koml\'{o}sy's stress-seeking verbs and stress avoiding verbs into obligatorily stressed verbs, potentially stressed verbs, obligatorily unstressed verbs and potentially unstressed verbs. 

Our classification of verbs into semantic classes is inspired by \cite{komlosy1989fokuszban}, since even though his classification is not motivated by semantics, there are apparent semantic similarities shared among verbs he groups together. For example, many stress-seeking verbs are experiencer-subject psych verbs and many stress-avoiding verbs express spatial configuration. 


\section{Methods}
\label{sec:methods}

\subsection{Data}
\label{ssec:data}
In this section, we describe how we build our dataset of verb-object pairs. Our data comes from the Hungarian Gigaword Corpus \citep{oravecz2014hungarian}, the largest, carefully curated multi-genre corpus of Hungarian containing 1.5 billion tokens. The distribution of tokens across genres and the source of the texts is summarized in Table~\ref{tab:gigaword}. The corpus is tokenized and morphologically analyzed via HunMorph \citep{tron2005hunmorph}, an FST-based parser that produces quite reliable annotations. 

\begin{table}[h]
\centering
\resizebox{\linewidth}{!}{
\begin{tabular}{@{}lrl@{}}
\toprule
\textbf{Genre}       & \textbf{\% of Tokens} & \textbf{Source}                         \\ \midrule
Journalism           & 42.0\%                          & Daily / weekly newspapers               \\
Personal             & 22.2\%                        & Social media                            \\
Literature           & 14.5\%                        & Digital Literary Academy                \\
Official             & 8.8\%                         & Documents from public admin.            \\
(Popular) science    & 7.2\%                         & Wikipedia, Hungarian Electronic Library \\
(Transcribed) spoken & 5.4\%                         & Radio programs                          \\ \bottomrule
\end{tabular}
}
\caption{Distribution of genres in the Hungarian Gigaword Corpus.}
\label{tab:gigaword}
\end{table}

\subsubsection{Dependency Parsing}
\label{sssec:dependency_parsing}

To obtain verb-object pairs, we perform dependency parsing on the data. We experiment with four different parsers: \emph{magyarlanc} \citep{zsibrita2013magyarlanc}, a widely used toolkit for lemmatization, dependency parsing and morphological analyses in Hungarian; Hungarian models for SpaCy\footnote{\url{https://github.com/oroszgy/spacy-hungarian-models}}; the Stanford NLP parser \citep{qi2018universal} and our own, rule-based parser. With the rule-based parser, we identify objects in the morphologically analyzed corpus by extracting nouns with accusative case that are only separated from the verb by adverb(s) and/or a preverb, or determiners and adjectives (if the noun is postverbal).

We randomly sample a list of 20 sentences from each genre in the corpus and manually compare the results of the four parsers. Specifically, we look at whether each parser accurately identifies verbs and their nominal objects. We find that the StanfordNLP parser and the rule-based parser are by far the best in terms of precision. However, the Stanford NLP parser outperforms the rule-based parser in terms of recall, which is expected, as it is able to identify longer-range dependencies. Given its performance, we used the Stanford NLP parser to extract verb-object pairs. Henceforth, we refer to each co-occurrence of a verb-object pair in the corpus as a \token.

\subsubsection{Morphological Analysis}
\label{sssec:morphological_analysis}
The StanfordNLP parser generates part-of-speech tags and morphological analyses for the words, based on tags that are supported by the Szeged Dependency Treebank \citep{vincze2010dependency}.\footnote{\url{https://universaldependencies.org/treebanks/hu\_szeged/index.html}} As mentioned in Section~\ref{ssec:obj_definiteness}, we use these definiteness tags on the verbs (\texttt{def} or \texttt{ind}) to determine the definiteness of the object. We find that the StanfordNLP parser detects definiteness marking on verbs with high accuracy, but unfortunately its lemmatization capability is very poor. Therefore, we re-lemmatize the verbs and objects using HunMorph \citep{tron2005hunmorph}, which often provides multiple possible lemmas. We use all possible lemmas (separated by a forward slash) to represent each verb and object. Henceforth, we refer to these combinations of possible lemmas simply as \lemmas{}.

\subsubsection{Filtering}

We perform multiple filtering steps to minimize the number of confounds in our data. Below we explain the goal and the details of each step. Before filtering, we start out with 17 million \tokens, representing 35,838 unique verb \lemmas{}, which includes verb \lemmas{} prefixed with preverbs.

\paragraph{Preverbs.} Preverbs occupy the preverbal slot, except when the sentence has contrastive focus, in which case they occur postverbally. Therefore, verbs with preverbs are constrained in their ordering with respect to their objects compared to other verbs. To eliminate the confound of preverbs having an influence on the ordering of verbs and their objects, we remove all verbs with preverbs. We detect verbs with preverbs via 1) morphological analysis, since preverbs can be prefixed to the verb, and 2) via dependency parsing (checking if there is a \texttt{compound:preverb} relation), in case the preverb is separated from the verb. Given the large number of preverb + verb combinations in Hungarian, this step results in the removal of 24,779 verb \lemmas{}, so 69\% of the \lemmas{} we started out with. At the end of this step, our dataset contains 11,059 unique verb \lemmas{}.

\paragraph{Bare objects.} Bare objects -- i.e. common nouns without modifiers -- also behave in a special way in Hungarian, as they are considered to be part of incorporating constructions (see Section~\ref{ssec:noun_incorporation}). The behavior of bare nouns is similar to preverbs, in that they always precede the verb in the absence of contrastive focus. Hence, we remove these objects from our analyses as well. We detect if a noun is bare by checking if it has zero dependents --- we only keep a noun with zero dependents if it is a proper noun. 

\paragraph{Transitivity.} To ensure the verbs are transitive, we keep those verbs that occur with objects in the corpus at least 30\% of the time. We acknowledge that this filtering step generates false negatives, as they might be transitive verbs that occur with overt objects less than 30\% of the time. However, it was more important for us to ensure high precision than recall --- i.e. to make sure that the verbs we find are indeed transitive. This filtering step results in the removal of 4391 verb \lemmas{}, with 6668 verb \lemmas{} remaining.

\paragraph{Multiple arguments.} We also wanted to make sure that our results are not skewed by verbs that frequently take more than one argument (e.g., ditransitive verbs and verbs with locative / directional arguments). The reason for this is that these verbs might still have ordering preferences, but not with respect to their objects but with respect to one of their other arguments. Such verbs would introduce a confound in our analyses, since by only looking at the ordering of the verb and the object, we would not be able to know if there is an additional argument influencing their relative ordering. Thus, we filter out those transitive verbs that occur with either of the following dependency relations more than 50\% of the time: \texttt{iobj}, \texttt{obl}, \texttt{nmod:obl}, \texttt{advmod:obl}, \texttt{amod:obl} and \texttt{ccomp:obl}.  As a result of this step, we remove 3034 \lemmas{} and have 3634 \lemmas{} remaining. 

In addition, we also remove all \tokens{} where there is an oblique argument beside the object, even if the \lemma{} does not occur with obliques more than 50\% of the time. This step is to ensure that for the \tokens{} we study, there is no additional argument influencing the verb-object ordering.

\paragraph{Frequency.} To ensure that we have a large enough number of \tokens{} for each verb for statistically robust analyses, we keep those verbs that occur at least 200 times in our data after the filtering steps. This way, even when we split the data into halves (Section~\ref{ssec:split}), both halves will have about 100 \tokens{} for each verb. These filtering steps yield 380 verb \lemmas{}, which can be found in Appendix~\ref{sec:appendix_verbs} along with their frequencies in our training data (Section~\ref{ssec:split}). Even though these verbs are all relatively frequent, there is still a discrepancy among their frequencies, as they follow a Zipfian distribution with a long tail (see Figure~\ref{fig:zipf} in the Appendix). For example, the most frequent verbs, \emph{ismer} `know (someone)', \emph{t\'amogat} `support' and \emph{okoz} `cause' are more than 20 times as frequent as the least frequent verbs \emph{pontoz} `score (a test)', \emph{aktiviz\'al} `get (someone) to take action', \emph{k\"orvonalaz} `outline' and \emph{mormol} `murmur'. 

\subsubsection{Splitting the Data}
\label{ssec:split}

Our final dataset includes approximately 1.3 million \tokens{} of verb-object pairs representing 380 verbs. To avoid overfitting to our data, we split it in two equal halves: a training set and test set. Both halves of the data contain at least 100 \tokens{} for each verb. Following standard practice in machine learning, we use the training set to develop our model and we apply our finalized model to the test set.


\subsection{Verb Classification}
\label{ssec:verb_classes}

In order to study the effect of lexical semantics on verb-object ordering, we classify verbs into coarse-grained semantic categories. Each semantic class is created because 1) it represents a highly salient lexical semantic category or 2) because it represents a semantic distinction that seems relevant to ordering preference. To meet condition 1), we consult previous literature on the semantic classification of verbs \citep[]{levin1993english}. To meet condition 2), we look at semantic distinctions between verbs with different ordering preferences in the training data to identify semantic features that clearly distinguish among verbs with different ordering preferences. It is important that we consider verbs' ordering preference only when we define semantic criteria for the verb classes, not when we assign the verbs to classes, so that we do not bias our experimental design.\footnote{Classifying verbs based on their ordering preferences would mean that we are not actually classifying them along the semantic definitions of classes we set up. This would clearly introduce bias and prevent us from meeting our research goal of studying the effect of verb semantics on verb-object ordering.}

 We define ten semantic classes, as well as a small \textsc{other} class for 18 verbs that are polysemous or cannot be assigned to any of the categories. The ten classes are: 
 \begin{enumerate}
     \item  \textsc{activity} (e.g., \emph{keres} `search for', \emph{firtat} `dwell on', \emph{foglalkoztat} `employ, occupy')
     \item \textsc{affect} (e.g., \emph{tiszt\'it} `clean', \emph{vereget} `hit at', \emph{s\"urget} `urge')
     \item \textsc{change} (e.g. \emph{aktiv\'al} `activate', \emph{\'erlel} `ripen', \emph{m\'elyit} `deepen, aggravate')
     \item \textsc{covering} (e.g. \emph{\"ovez} `surround', \emph{\'ov} `guard', \emph{tartalmaz} `contain)
     \item \textsc{creation/representation} (e.g. \emph{alkot} `create', \emph{szeml\'eltet} `illustrate', \emph{szapor\'it} `breed')
     \item \textsc{evaluation/experience} (e.g., \emph{gy\H{u}l\"ol} `hate', \emph{csod\'al} `admire', \emph{un} `be bored by')
     \item \textsc{ingestion} (e.g., \emph{fogyaszt} `consume', \emph{fal} `devour', \emph{kortyol} `take sips of')
     \item \textsc{ownership} (e.g., \emph{birtokol} `own, possess', \emph{\'erdemel} `deserve', \emph{illet} `belong to')
     \item \textsc{perception} (e.g., \emph{hall} `hear', \emph{vizsg\'al} `examine', \emph{\'erzekel} `perceive')
     \item \textsc{preference} (e.g., \emph{prefer\'al} `prefer', \emph{v\'alaszt} `choose', \emph{latolgat} `ponder on (a decision/choice)')
     \item ? (\textsc{other}) (e.g.  \emph{szerkeszt} `edit', \emph{d\'edelget} `fondle, pamper', \emph{hallat} `make heard')
 \end{enumerate}
 
 For most of our verb classes, both conditions 1) and 2) are met. There are two exceptions where one of these conditions factored much more prominently into our decision to create the verb class than the other. The first one is the separation of change of state verbs (\textsc{change}) from verbs implying force exertion (\textsc{affect}) --- here, condition 1) holds more strongly than condition 2) \cite[see][p. 240]{levin1993english}. The second exception is verbs implying preference (\textsc{preference}), where condition 2) is met, but condition 1) less so.

 Figure~\ref{fig:flowchart} shows the classes and the decision procedure we used to assign verbs to classes.  The series of questions that lead to each category in the flowchart constitutes the definition of each category. We color code each verb class in the flowchart for their ordering preference, which is based on patterns observed in the training data. We discuss the ordering preference of each verb class in Section~\ref{sec:results}.

\newgeometry{scale=1}
  {\thispagestyle{empty}
  \centering
  \includegraphics[scale=.95]{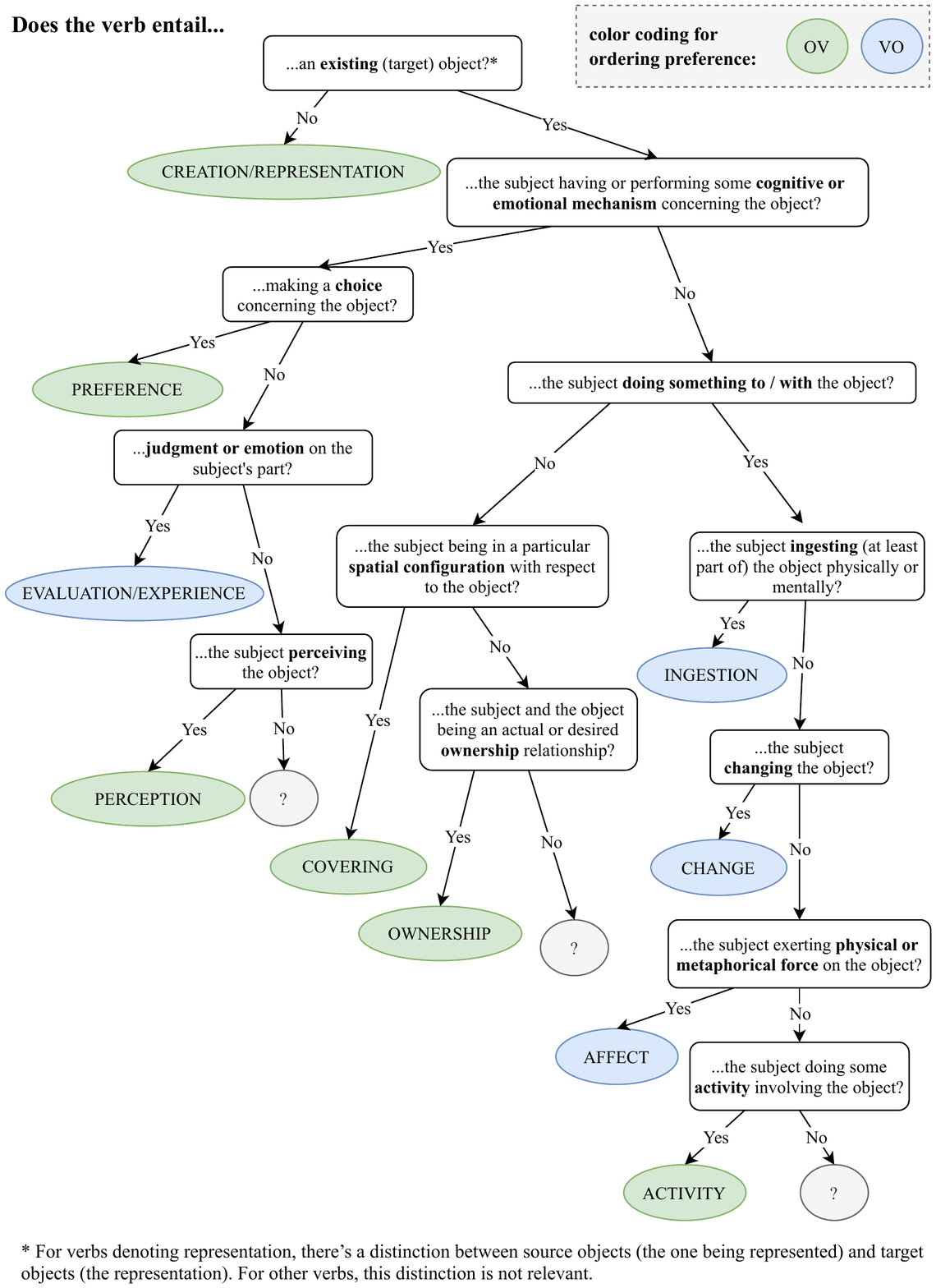}
  \setlength{\abovecaptionskip}{-15pt}
  \captionof{figure}{Flowchart illustrating our verb classification procedure. }
  \label{fig:flowchart}}
\restoregeometry
\clearpage
\newpage
\doublespacing

\subsection{Object Features}
\label{ssec:object_features}

In Section~\ref{sec:intro}, we motivated our use of two control features, object definiteness and object NP weight, which we compare with the effect of the verb's semantic class.

\paragraph{Object definiteness.} In Section~\ref{ssec:obj_definiteness}, we motivated our use of object definiteness as a feature and provided a brief overview of definiteness marking in Hungarian. We detect the definiteness of the object as part of the morphological analysis (Section~\ref{sssec:morphological_analysis}), by looking at the definiteness marking on verbs.

\paragraph{Object NP weight.} The complexity of constituents can be predictive of their relative ordering, given that ``constituents in many languages tend to occur in increasing size and complexity''\cite[][p. 81]{wasow1997remarks}. Complexity and size can be calculated in different ways. \citet{bresnan2007predicting} uses length of the constituents as a predictor of their relative ordering in the dative alternation and \citet{benor2006chicken} uses the number of syllables in each noun as a predictor of their ordering in binomial expressions. We use the weight of the object NP (i.e. the number of elements in it) as an estimate of its complexity. We estimate object NP weight using the StanfordNLP parser \citep{qi2018universal}, by calculating the number of dependents of the object's head noun.

\subsection{Logistic Regression}
\label{ssec:model}

We use a simple logistic regression model to see how well our features can predict the ordering of verbs and their objects. The binary response variable is the ordering of verbs and their objects (0 for OV, 1 for VO). Our set of predictors include categorical variables -- the verb's lemma, the verb's semantic class, the definiteness of the object -- and a continuous variable -- the weight of the object NP. The model estimates the probability $p$ of a verb preceding the object via

\begin{equation}
   p= \dfrac{1}{1 + e^{(\beta_0 + \beta_1x_1 + ... + \beta_nx_n)}}
   \label{eq:logistic_regression}
\end{equation}

\noindent where the $\beta_i$ is the weight (or parameter) associated with variable $x_i$. As for categorical variables, each verb category is assigned a separate binary variable, except for the alphabetically first category (in our case, the \textsc{other} category represented by ?), which is estimated by the intercept ($\beta_0$). Each continuous variable is assigned to a single continuous variable.


\subsubsection{Estimating Ordering Preference}

We estimate the ordering preference of each verb by observing if it is significantly more likely to precede or follow its objects in our data. To this end, we build a logistic regression model with a single predictor, the verb's lemma, and estimate the parameters for each verb. The model estimates the log odds of a verb preceding or following its object. If the log odds is negative, then the verb is more likely to precede its object, and if the log odds if positive, then it is more likely to follow its object. 

We determine the significance of the estimate for each verb by $z$-scoring the log odds (dividing the log odds by the standard error). We consider the ordering preference of a verb to be significant if the absolute value of the $z$-score exceeds 2, which means that the estimate is greater than 2 standard deviations. We follow the same procedure for determining the ordering preference of each semantic class, by running a logistic regression model with verb class as a single predictor.

\subsubsection{Model Accuracy}

We treat the accuracy of each model as an estimate for how well the features in that model can explain verb-object ordering in our data. Accuracy is defined as the proportion of \tokens{} whose ordering the model predicts correctly. To see whether our models are overfitting to the training set, we also use our models (trained on the training set) to predict the ordering of \tokens{} in the test set. If the difference between the training and test accuracies is not significant (as measured by a two-sample t-test on the training vs test predictions), that means that our parameter estimates are not biased towards the training set.

\section{Results \& Discussion}
\label{sec:results}

In this section, we address each of the research questions outlined in Section~\ref{sec:research_questions}.

\subsection{Ordering Preferences of Verbs}
We find that 280 out of 380 verbs (74\%) have a significant ordering preference. Out of the verbs with a significant ordering preference, 106 (38\%) have an OV preference and 174 (62\%) have a VO preference.
The fact that more verbs have a VO preference than an OV preference shows that focusing objects (resulting in an OV order) is less frequent across verbs than not focusing them. It is interesting that OV-preferring verbs still do make up a significant proportion of verbs with an ordering preference (38\%). These OV-preferring verbs are transitive verbs that tend to occur with focused objects, and they form a larger group than has been identified in previous work by \citet{komlosy1989fokuszban}. The ordering preference of all verbs in our data is included in Appendix~\ref{sec:appendix_verbs}. 

In our first research question, we also asked how well we can predict the ordering of verbs and objects in our data based on the verb exclusively. We sought to answer this question using a model that only includes the lemma of the verb as a feature --- we call this model the \textsc{verb-only} model.  We found that the accuracy of this model is 68\%, as shown in Table~\ref{tab:results}, with no significant difference between the training and the test set ($p \approx 0.3$, obtained via a two sample t-test).\footnote{It is important to note that the upper bound on classification accuracy is lower than 100\%, as there is also some amount of free variation in verb-object ordering in Hungarian \citep{kiss1994scrambling}. We leave the estimation of the amount of this free variation for future work.}  We compared the accuracy of the \textsc{verb-only} model with the majority baseline that randomly predicts VO ordering with the same frequency as observed in the training data (53\%). The accuracy of the \textsc{verb-only} model is 1.3 times better than the majority baseline with very high significance ($p < 0.001$). This result suggests that the verb does explain a significant portion of the variance in verb-object order.

 \begin{table}[]
 \centering
\begin{tabular}{@{}lcc@{}}
\toprule
\textbf{Model}                                                                     & \textbf{Train Acc.} & \textbf{Test Acc.} \\ \midrule
majority baseline                                                                  & 52.91\%        & 53.04\%       \\
\textsc{verb-only}                                                                         & 67.95\%        & 67.82\%       \\
\textsc{class-only}                                                                         & 65.04\%        & 64.93\%       \\
\textsc{object-weight-only}                                                                  & 54.90\%        & 54.99\%       \\
\textsc{definiteness-only}                                                             & 57.29\%        & 57.17\%       \\
\begin{tabular}[c]{@{}l@{}}\textsc{complex}\\ (class+object-size+definiteness)\end{tabular} & 67.63\%        & 67.52\%       \\ \bottomrule
\end{tabular}
\caption{Accuracy of logistic regression models, using different features, on the training and the test sets.}
\label{tab:results}
\end{table}

\subsection{Ordering Preferences of Verb Classes}
In Section~\ref{ssec:verb_classes}, we described our verb classification procedure; this procedure included identifying a small set of semantic classes for the verbs, which are salient based on previous literature and based on the semantic features that are relevant for the verbs' ordering preference. Below we summarize our high-level findings regarding the relationship between the semantic verb classes and their ordering preference.

\begin{itemize}
    \item Stative verbs tend to be OV-preferring, especially if they denote location or spatial configuration. Both experiencer-subject and experiencer-object psych verbs are exceptions among stative verbs, as they are nearly all VO-preferring.
    \item Non-stative verbs, especially ones that entail the subject doing something to or with an existing object, tend to be VO-preferring. Verbs of creation / representation, which do not entail an existing target object,\footnote{As the footnote to Figure~\ref{fig:flowchart} mentions, for verbs denoting representation, there’s a distinction between source objects (the one being represented) and target objects (the representation). For other verbs, this distinction is not relevant.} are exceptions to this, as they tend to be OV-preferring.
\end{itemize}

The verb classes are listed in Table~\ref{tab:verb_classes}, along with their ordering preference and example class members showing a significant ordering preference. The largest OV-preferring class is \textsc{creation/representation} with 50 verbs,  and the largest VO-preferring class is \textsc{change}, with 110 verbs. The smallest OV-preferring class is \textsc{perception}, with 6 verbs, and the smallest VO-preferring class is \textsc{ingestion}, with 11 verbs. The disparity between the class sizes can be partially explained by the disparity between the semantic granularity of the classes --- for example, \textsc{change} is a much broader category semantically than \textsc{ingestion}. In Section~\ref{ssec:verb_classes}, we identified an \textsc{other} class of 18 verbs (5\%) that were either highly polysemous or they did not fit the definition of any class; this class, which included 18 verbs, was included in all our models.

\begin{table}[]
\resizebox{\textwidth}{!}{%
\begin{tabular}{|l|c|c|c|c|l|l|}
\hline
\multicolumn{1}{|c|}{\multirow{2}{*}{\textbf{Category}}}           & \multirow{2}{*}{\textbf{Pref.}} & \multicolumn{3}{c|}{\textbf{Verb Counts}}                                                    & \multicolumn{2}{c|}{\textbf{Example Verbs}}                                                                                                                                                                 \\ \cline{3-7} 
\multicolumn{1}{|c|}{}                                             &                                 & \textbf{OV} & \textbf{VO} & \textbf{\begin{tabular}[c]{@{}c@{}}No sig.\\ pref.\end{tabular}} & \multicolumn{1}{c|}{\textbf{OV}}                                                                    & \multicolumn{1}{c|}{\textbf{VO}}                                                                      \\ \hline
ACTIVITY                                                           & OV                              & 5           & 0           & 3                                                                & \begin{tabular}[c]{@{}l@{}}alkalmaz `employ', kutat `research',\\ üzemeltet `operate'\end{tabular}  &                                                                                                       \\ \hline
COVERING                                                           & OV                              & 13          & 2           & 4                                                                & \begin{tabular}[c]{@{}l@{}}fed `cover', díszít `decorate',\\ övez `surround'\end{tabular}           & óv `protect', kerülget `go around'                                                                    \\ \hline
\begin{tabular}[c]{@{}l@{}}CREATION/\\ REPRESENTATION\end{tabular} & OV                              & 35          & 2           & 13                                                               & \begin{tabular}[c]{@{}l@{}}alkot `create', tükröz `mirror', \\ jelent `mean'\end{tabular}           & szaporít `breed', vázol `sketch'                                                                      \\ \hline
OWNERSHIP                                                          & OV                              & 4           & 0           & 0                                                                & \begin{tabular}[c]{@{}l@{}}birtokol `possess', érdemel `deserve',\\ illet `belong to'\end{tabular}  &                                                                                                       \\ \hline
PERCEPTION                                                         & OV                              & 3           & 0           & 3                                                                & hall `hear', vizsgál `examine'                                                                      &                                                                                                       \\ \hline
PREFERENCE                                                         & OV                              & 4           & 0           & 1                                                                & \begin{tabular}[c]{@{}l@{}}választ `choose', preferál `prefer',\\ céloz `aim at'\end{tabular}       &                                                                                                       \\ \hline
AFFECT                                                             & VO                              & 14          & 47          & 30                                                               & \begin{tabular}[c]{@{}l@{}}csókol `kiss', fenyeget `threaten',\\ sürget `urge'\end{tabular}         & \begin{tabular}[c]{@{}l@{}}hajszol `rush',  bántalmaz `hurt, abuse',\\ támogat `support'\end{tabular} \\ \hline
CHANGE                                                             & VO                              & 12          & 74          & 24                                                               & \begin{tabular}[c]{@{}l@{}}ihlet `inspire', motivál `motivate',\\ tömörít `compactify'\end{tabular} & \begin{tabular}[c]{@{}l@{}}aktivál `activate', csökkent `reduce',\\ javít `repair'\end{tabular}       \\ \hline
\begin{tabular}[c]{@{}l@{}}EVALUATION/\\ EXPERIENCE\end{tabular}   & VO                              & 2           & 39          & 15                                                               & gyászol `mourn', hibáztat `blame'                                                                   & \begin{tabular}[c]{@{}l@{}}bírál `judge', csodál `admire',\\ gyűlöl `hate'\end{tabular}               \\ \hline
INGESTION                                                          & VO                              & 2           & 9           & 0                                                                & \begin{tabular}[c]{@{}l@{}}fogyaszt `consume',\\ vedel `drink (a lot of)'\end{tabular}              & \begin{tabular}[c]{@{}l@{}}fal `devour', iszik `drink',\\ olvas `read'\end{tabular}                   \\ \hline
OTHER                                                              & OV                              & 10          & 1           & 7                                                                & szerkeszt `edit', szolgál `serve'                                                                   & viszonoz `requite'                                                                                    \\ \hline
\end{tabular}%
}
\caption{Verb classes, their ordering preference (all of them are significant) and example verbs.}
\label{tab:verb_classes}
\end{table}

In this research question, we also asked how well we can predict verb-object ordering based on the verb's semantic class. To this end, we construct a \textsc{class-only} model, which achieves an accuracy of 65\%, with no significant difference between the training and test sets ($p<0.001$). This result supports the way in which we categorized the verbs, as we find that the ordering preference of our verb classes largely account for the ordering preference of individual verbs. Specifically, the accuracy of the \textsc{verb-only} model (an upper bound on verb classification quality) is 68\%, only 3\% higher than the accuracy of \textsc{class-only} model. This small difference in model accuracy is remarkable given that there is a large, 97\% reduction in the size of the feature set from the \textsc{verb-only} model (380 verbs) to the \textsc{class-only} model (11 classes). 


\subsection{Importance of Object-Related Features}

We construct a \textsc{definiteness-only} and an \textsc{object-weight-only} model to estimate the effect size of object definiteness and object NP weight, respectively. The \textsc{definiteness-only} model achieves an accuracy of 57\%  and the \textsc{object-weight-only} model achieves an accuracy of 55\% (training and test performance are the same for both models, $p < 0.001$). These results indicate that these object-related features are not as important as features related to the verb (verb lemma and verb class). 

Our \textsc{complex} model, which includes verb class, object definiteness and object NP weight as features, achieves 68\% accuracy, with no significant difference between training and test performance ($p < 0.001$) Thus, using the \textsc{complex} model with only 14 features (11 for verb class + 2 for object definiteness + 1 continuous variable for object NP weight), we can predict verb-object ordering with the same level of accuracy as using the \textsc{verb-only} model with 380 features. The fact that there is no significant difference between training and test accuracy for any of the models shows that the parameter estimates are not overfitting to the training set.

\section{Conclusion}
\label{sec:conclusion}

In this paper, we investigated the role of three features, including the verb's semantic class, object definiteness and object NP weight, in determining verb-object order in Hungarian. We extract verb-object pairs and their associated object-related features from the Hungarian Gigaword Corpus, manually assign verbs to semantic classes,  and build a logistic regression model to estimate the importance of different features. We discover patterns of semantic similarity among verbs with similar ordering preference, and we perform our verb classification based on these patterns.

We find that all of our features obtain significantly higher accuracy than the majority baseline in predicting verb-object ordering. Even though using the verb lemma as a feature seems to be the most effective among our features, the complexity of the \textsc{verb-only} model (with 380 features) is significantly greater than that of our \textsc{complex} model, which contains all other features (14 features total). The \textsc{verb-only} model and \textsc{complex} models obtain the same accuracy, 68\%, on both the training and the test sets. A \textsc{class-only} model achieves 65\% accuracy, only 3\% less than the \textsc{verb-only} model, suggesting that our semantic classification approximates the similarities among verbs in terms of ordering preference quite well. 

The predictive power of our lexical semantic feature shows that it might be playing a non-negligible role in Hungarian word order. We hope that this investigation will lead to more studies in this domain. Promising avenues for future work include extending our analysis to a larger number of verbs (e.g., verbs with multiple arguments), adding additional features (e.g., the animacy or humanness of arguments) and better understanding the interaction among discourse context, lexical semantic factors and perhaps other factors, such as cognitive constraints, that may influence word order in Hungarian.

\section{Acknowledgments}

We thank Andr\'as Koml\'osy for enlightening conversations. We are also grateful for the support of the Melvin and Joan Lane Stanford Graduate Fellowship to Dorottya Demszky.

\bibliography{main}
\bibliographystyle{apacite}

\appendix

\section{Verbs}
\label{sec:appendix_verbs}

The table below contains information about each verb lemma we study, including frequency and ordering preference and semantic class (Section~\ref{ssec:verb_classes}).

\footnotesize
\begin{longtable}{|llccccc|}
\hline 
\hline
\textbf{Verb Lemma}  & \textbf{Translation} & \textbf{Freq.}  & \textbf{Pref.} & \textbf{Log Odds} & \textbf{$z$-score} & \textbf{Category} \\ \hline
\endfirsthead
\endhead
aggaszt	&	worry	&	444	&	OV	&	-0.81	&	-7.85	&	\textsc{change}	\\ \hline
\begin{tabular}[c]{@{}l@{}}akadályoz/\\ akadályozik\end{tabular}	&	obstruct	&	3031	&	VO	&	0.89	&	8.08	&	\textsc{affect}	\\ \hline
aktivál	&	activate	&	488	&	VO	&	0.76	&	5.41	&	\textsc{change}	\\ \hline
aktivizál	&	\begin{tabular}[c]{@{}l@{}}get someone to\\ take action\end{tabular}	&	100	&	VO	&	1.32	&	4.98	&	\textsc{change}	\\ \hline
alkalmaz	&	\begin{tabular}[c]{@{}l@{}}employ,\\ implement\end{tabular}	&	888	&	OV	&	-1.6	&	-11.74	&	\textsc{activity}	\\ \hline
alkot	&	create, form	&	9081	&	OV	&	-1.08	&	-10.24	&	\begin{tabular}[c]{@{}c@{}}\textsc{creation/}\\\textsc{representation}\end{tabular}	\\ \hline
áramtalanít	&	de-electrify	&	122	&	no sig.	&	0.4	&	1.89	&	\textsc{change}	\\ \hline
árnyal	&	\begin{tabular}[c]{@{}l@{}}make nuanced,\\ complicate\end{tabular}	&	369	&	VO	&	0.54	&	3.61	&	\textsc{change}	\\ \hline
\begin{tabular}[c]{@{}l@{}}bántalmaz/\\ bántalmazik\end{tabular}	&	hurt, abuse	&	1165	&	VO	&	0.52	&	4.34	&	\textsc{affect}	\\ \hline
befolyásol	&	influence	&	9720	&	VO	&	0.46	&	4.39	&	\textsc{affect}	\\ \hline
bírál	&	judge, evaluate	&	3879	&	VO	&	0.95	&	8.71	&	\begin{tabular}[c]{@{}c@{}}\textsc{evaluation/}\\\textsc{experience}\end{tabular}	\\ \hline
birtokol	&	own, possess	&	1309	&	OV	&	-0.53	&	-4.51	&	\textsc{ownership}	\\ \hline
blokkol	&	block	&	417	&	VO	&	0.71	&	4.88	&	\textsc{affect}	\\ \hline
bojkottál	&	boycott	&	525	&	VO	&	1.08	&	7.53	&	\textsc{affect}	\\ \hline
boncolgat	&	\begin{tabular}[c]{@{}l@{}}anatomize\\ (iterative)\end{tabular}	&	231	&	OV	&	-1.66	&	-8.02	&	\textsc{change}	\\ \hline
bontogat	&	unfold (iterative)	&	116	&	VO	&	0.57	&	2.59	&	\textsc{change}	\\ \hline
bonyolít	&	complicate	&	1090	&	OV	&	-0.26	&	-2.16	&	\textsc{change}	\\ \hline
borít	&	cover	&	2486	&	OV	&	-1.21	&	-10.68	&	\textsc{covering}	\\ \hline
cáfol	&	refute	&	2643	&	VO	&	1.32	&	11.69	&	\begin{tabular}[c]{@{}c@{}}\textsc{evaluation/}\\\textsc{experience}\end{tabular}	\\ \hline
\begin{tabular}[c]{@{}l@{}}céloz/\\ célozik\end{tabular}	&	aim at	&	1465	&	OV	&	-3.57	&	-18.76	&	\textsc{preference}	\\ \hline
\begin{tabular}[c]{@{}l@{}}cenzúráz/\\ cenzúrázik\end{tabular}	&	censor	&	128	&	no sig.	&	0.13	&	0.61	&	\textsc{affect}	\\ \hline
csapkod	&	hit at	&	260	&	VO	&	0.47	&	2.87	&	\textsc{affect}	\\ \hline
cserélget	&	switch, replace	&	134	&	VO	&	1.29	&	5.51	&	\textsc{change}	\\ \hline
csillapít	&	soothe	&	438	&	VO	&	0.65	&	4.53	&	\textsc{change}	\\ \hline
csodál	&	admire	&	676	&	VO	&	0.89	&	6.67	&	\begin{tabular}[c]{@{}c@{}}\textsc{evaluation/}\\\textsc{experience}\end{tabular}	\\ \hline
csökkent	&	reduce	&	8782	&	VO	&	1.3	&	12.29	&	\textsc{change}	\\ \hline
csókol	&	kiss	&	1013	&	OV	&	-1.22	&	-9.59	&	\textsc{affect}	\\ \hline
csókolgat	&	kiss	&	144	&	no sig.	&	-0.03	&	-0.14	&	\textsc{affect}	\\ \hline
csorbít	&	impair	&	353	&	VO	&	0.47	&	3.11	&	\textsc{change}	\\ \hline
csóvál	&	frisk	&	855	&	VO	&	0.36	&	2.92	&	\textsc{affect}	\\ \hline
dédelget	&	fondle, pamper	&	174	&	OV	&	-0.94	&	-4.74	&	\textsc{other}\\ \hline
dicsér	&	praise	&	2163	&	no sig.	&	0.04	&	0.37	&	\begin{tabular}[c]{@{}c@{}}\textsc{evaluation/}\\\textsc{experience}\end{tabular}	\\ \hline
dicsőít	&	glorify	&	231	&	no sig.	&	0.25	&	1.5	&	\begin{tabular}[c]{@{}c@{}}\textsc{evaluation/}\\\textsc{experience}\end{tabular}	\\ \hline
diktál	&	dictate	&	645	&	no sig.	&	-0.18	&	-1.42	&	\begin{tabular}[c]{@{}c@{}}\textsc{creation/}\\\textsc{representation}\end{tabular}	\\ \hline
díszít	&	embellish	&	1634	&	OV	&	-1.49	&	-12.3	&	\textsc{covering}	\\ \hline
diszkriminál	&	discriminate	&	120	&	VO	&	0.93	&	4.09	&	\textsc{affect}	\\ \hline
dokumentál	&	document	&	327	&	no sig.	&	-0.02	&	-0.12	&	\begin{tabular}[c]{@{}c@{}}\textsc{creation/}\\\textsc{representation}\end{tabular}	\\ \hline
dönget	&	\begin{tabular}[c]{@{}l@{}}pound on,\\ hit on\end{tabular}	&	248	&	OV	&	-0.65	&	-3.86	&	\textsc{affect}	\\ \hline
dörzsöl	&	rub	&	378	&	VO	&	0.61	&	4.1	&	\textsc{affect}	\\ \hline
drágít	&	\begin{tabular}[c]{@{}l@{}}make more\\ expensive\end{tabular}	&	123	&	VO	&	1.83	&	6.52	&	\textsc{change}	\\ \hline
dúdol	&	hum	&	108	&	no sig.	&	0.15	&	0.68	&	\begin{tabular}[c]{@{}c@{}}\textsc{creation/}\\\textsc{representation}\end{tabular}	\\ \hline
éget	&	burn	&	548	&	no sig.	&	0.23	&	1.75	&	\textsc{change}	\\ \hline
egyenget	&	smooth	&	263	&	no sig.	&	0.08	&	0.52	&	\textsc{change}	\\ \hline
egységesít	&	unify	&	242	&	VO	&	1.18	&	6.43	&	\textsc{change}	\\ \hline
egyszerűsít	&	simplify	&	423	&	VO	&	1.38	&	8.68	&	\textsc{change}	\\ \hline
ékesít	&	decorate	&	146	&	OV	&	-1.08	&	-5	&	\textsc{covering}	\\ \hline
éltet	&	\begin{tabular}[c]{@{}l@{}}keep alive,\\ sustain\end{tabular}	&	649	&	no sig.	&	0.23	&	1.74	&	\textsc{affect}	\\ \hline
elemez	&	analyze	&	1137	&	no sig.	&	-0.03	&	-0.28	&	\begin{tabular}[c]{@{}c@{}}\textsc{evaluation/}\\\textsc{experience}\end{tabular}	\\ \hline
\begin{tabular}[c]{@{}l@{}}elemez/\\ elemezik\end{tabular}	&	analyze	&	1384	&	no sig.	&	0.04	&	0.37	&	\begin{tabular}[c]{@{}c@{}}\textsc{evaluation/}\\\textsc{experience}\end{tabular}	\\ \hline
élénkít	&	vitalize	&	214	&	VO	&	1.16	&	6.1	&	\textsc{change}	\\ \hline
ellenez	&	oppose	&	1794	&	VO	&	1.59	&	13.22	&	\begin{tabular}[c]{@{}c@{}}\textsc{evaluation/}\\\textsc{experience}\end{tabular}	\\ \hline
ellenőriz	&	check, verify	&	2988	&	VO	&	0.37	&	3.34	&	\begin{tabular}[c]{@{}c@{}}\textsc{evaluation/}\\\textsc{experience}\end{tabular}	\\ \hline
\begin{tabular}[c]{@{}l@{}}ellensúlyoz/\\ ellensúlyozik\end{tabular}	&	counterbalance	&	709	&	no sig.	&	0.05	&	0.42	&	\textsc{covering}	\\ \hline
élvez/élvezik	&	enjoy	&	6224	&	VO	&	0.8	&	7.49	&	\begin{tabular}[c]{@{}c@{}}\textsc{evaluation/}\\\textsc{experience}\end{tabular}	\\ \hline
emelget	&	lift	&	212	&	VO	&	0.73	&	4.07	&	\textsc{affect}	\\ \hline
enyhít	&	ease	&	1531	&	VO	&	1	&	8.51	&	\textsc{change}	\\ \hline
érdemel	&	deserve	&	1254	&	OV	&	-1.43	&	-11.39	&	\textsc{ownership}	\\ \hline
eredményez	&	result in	&	1416	&	OV	&	-2.75	&	-18.11	&	\begin{tabular}[c]{@{}c@{}}\textsc{creation/}\\\textsc{representation}\end{tabular}	\\ \hline
\begin{tabular}[c]{@{}l@{}}eredményez/\\ eredményezik\end{tabular}	&	result in	&	4176	&	OV	&	-2.13	&	-18.64	&	\begin{tabular}[c]{@{}c@{}}\textsc{creation/}\\\textsc{representation}\end{tabular}	\\ \hline
érint	&	touch (on)	&	17011	&	OV	&	-0.81	&	-7.8	&	\textsc{affect}	\\ \hline
érlel	&	ripen	&	126	&	no sig.	&	-0.19	&	-0.93	&	\textsc{change}	\\ \hline
erőltet	&	force	&	674	&	VO	&	0.86	&	6.44	&	\textsc{affect}	\\ \hline
erősít	&	strengthen	&	5690	&	no sig.	&	0.03	&	0.27	&	\textsc{change}	\\ \hline
\begin{tabular}[c]{@{}l@{}}értelmez/\\ értelmezik\end{tabular}	&	\begin{tabular}[c]{@{}l@{}}interpret,\\ make sense of\end{tabular}	&	1831	&	VO	&	0.45	&	3.93	&	\begin{tabular}[c]{@{}c@{}}\textsc{evaluation/}\\\textsc{experience}\end{tabular}	\\ \hline
érvénytelenít	&	void	&	460	&	VO	&	0.78	&	5.4	&	\textsc{change}	\\ \hline
érzékel	&	perceive	&	1836	&	no sig.	&	0.19	&	1.66	&	\textsc{perception}	\\ \hline
érzékeltet	&	make perceive	&	631	&	no sig.	&	0.07	&	0.51	&	\textsc{other}\\ \hline
fájlal	&	\begin{tabular}[c]{@{}l@{}}feel/express\\ pain\end{tabular}	&	218	&	no sig.	&	-0.17	&	-0.97	&	\begin{tabular}[c]{@{}c@{}}\textsc{evaluation/}\\\textsc{experience}\end{tabular}	\\ \hline
fal	&	devour	&	262	&	VO	&	1.44	&	7.69	&	\textsc{ingestion}	\\ \hline
favorizál	&	favor	&	144	&	OV	&	-1.42	&	-6.07	&	\textsc{preference}	\\ \hline
fed	&	cover	&	1404	&	OV	&	-0.42	&	-3.58	&	\textsc{covering}	\\ \hline
fejleszt	&	develop	&	2691	&	no sig.	&	0.08	&	0.74	&	\begin{tabular}[c]{@{}c@{}}\textsc{creation/}\\\textsc{representation}\end{tabular}	\\ \hline
fékez/fékezik	&	put a halt to, break	&	425	&	VO	&	0.93	&	6.26	&	\textsc{affect}	\\ \hline
félt	&	\begin{tabular}[c]{@{}l@{}}worry about,\\ be protective of\end{tabular}	&	1377	&	VO	&	0.67	&	5.73	&	\begin{tabular}[c]{@{}c@{}}\textsc{evaluation/}\\\textsc{experience}\end{tabular}	\\ \hline
fémjelez	&	mark (a metal)	&	166	&	OV	&	-0.42	&	-2.2	&	\textsc{change}	\\ \hline
fen	&	hone	&	107	&	VO	&	1.89	&	6.22	&	\textsc{change}	\\ \hline
fenyeget	&	threaten	&	3373	&	OV	&	-0.59	&	-5.44	&	\textsc{affect}	\\ \hline
\begin{tabular}[c]{@{}l@{}}fényképez/\\ fényképezik\end{tabular}	&	photograph	&	286	&	no sig.	&	-0.24	&	-1.52	&	\begin{tabular}[c]{@{}c@{}}\textsc{creation/}\\\textsc{representation}\end{tabular}	\\ \hline
fertőtlenít	&	disinfect	&	178	&	no sig.	&	0.23	&	1.24	&	\textsc{change}	\\ \hline
feszeget	&	\begin{tabular}[c]{@{}l@{}}literal: stretch;\\ figurative: discuss,\\ stretch the\\boundaries of\end{tabular}	&	640	&	OV	&	-0.87	&	-6.47	&	\textsc{other}\\ \hline
finomít	&	refine	&	164	&	VO	&	0.68	&	3.51	&	\textsc{change}	\\ \hline
firtat	&	dwell on	&	403	&	OV	&	-1.19	&	-7.61	&	\textsc{activity}	\\ \hline
foglalkoztat	&	employ, occupy	&	2046	&	OV	&	-1.16	&	-10.06	&	\textsc{activity}	\\ \hline
fogyaszt	&	consume	&	3163	&	OV	&	-0.91	&	-8.23	&	\textsc{ingestion}	\\ \hline
fojtogat	&	choke (iterative)	&	143	&	no sig.	&	0.15	&	0.78	&	\textsc{affect}	\\ \hline
fokoz/fokozik	&	intensify, elevate	&	3262	&	VO	&	0.61	&	5.62	&	\textsc{change}	\\ \hline
fontolgat	&	\begin{tabular}[c]{@{}l@{}}ponder upon\\ a decision\end{tabular}	&	835	&	OV	&	-0.79	&	-6.25	&	\textsc{preference}	\\ \hline
forradalmasít	&	revolutionize	&	223	&	VO	&	1.65	&	7.88	&	\textsc{change}	\\ \hline
fosztogat	&	loot (iterative)	&	147	&	no sig.	&	-0.04	&	-0.21	&	\textsc{affect}	\\ \hline
frissít	&	refresh	&	654	&	VO	&	0.73	&	5.55	&	\textsc{change}	\\ \hline
fürkész	&	scan, examine	&	211	&	no sig.	&	-0.28	&	-1.6	&	\textsc{perception}	\\ \hline
fűt	&	heat	&	333	&	no sig.	&	0.04	&	0.28	&	\textsc{change}	\\ \hline
garantál	&	guarantee	&	1618	&	VO	&	0.48	&	4.19	&	\begin{tabular}[c]{@{}c@{}}\textsc{evaluation/}\\\textsc{experience}\end{tabular}	\\ \hline
gátol	&	obstruct	&	1761	&	VO	&	0.83	&	7.25	&	\textsc{affect}	\\ \hline
generál	&	generate	&	679	&	OV	&	-1.33	&	-9.54	&	\begin{tabular}[c]{@{}c@{}}\textsc{creation/}\\\textsc{representation}\end{tabular}	\\ \hline
gerjeszt	&	induce	&	571	&	OV	&	-0.7	&	-5.18	&	\begin{tabular}[c]{@{}c@{}}\textsc{creation/}\\\textsc{representation}\end{tabular}	\\ \hline
\begin{tabular}[c]{@{}l@{}}gondoz/\\ gondozik\end{tabular}	&	care for	&	714	&	no sig.	&	-0.06	&	-0.49	&	\textsc{affect}	\\ \hline
gúnyol	&	mock	&	119	&	no sig.	&	0.19	&	0.88	&	\begin{tabular}[c]{@{}c@{}}\textsc{evaluation/}\\\textsc{experience}\end{tabular}	\\ \hline
gyaláz	&	revile	&	210	&	no sig.	&	0.17	&	1	&	\begin{tabular}[c]{@{}c@{}}\textsc{evaluation/}\\\textsc{experience}\end{tabular}	\\ \hline
gyarapít	&	breed	&	717	&	OV	&	-0.41	&	-3.21	&	\begin{tabular}[c]{@{}c@{}}\textsc{creation/}\\\textsc{representation}\end{tabular}	\\ \hline
gyászol	&	mourn	&	367	&	OV	&	-0.55	&	-3.7	&	\begin{tabular}[c]{@{}c@{}}\textsc{evaluation/}\\\textsc{experience}\end{tabular}	\\ \hline
gyengít	&	weaken	&	1024	&	VO	&	0.82	&	6.63	&	\textsc{change}	\\ \hline
gyógyít	&	heal	&	500	&	no sig.	&	0.15	&	1.12	&	\textsc{change}	\\ \hline
gyorsít	&	accelerate	&	639	&	VO	&	1.3	&	9.22	&	\textsc{change}	\\ \hline
gyűlöl	&	hate	&	1330	&	VO	&	1.74	&	13.58	&	\begin{tabular}[c]{@{}c@{}}\textsc{evaluation/}\\\textsc{experience}\end{tabular}	\\ \hline
habzsol	&	devour	&	121	&	VO	&	2.52	&	6.98	&	\textsc{ingestion}	\\ \hline
hajszol	&	rush	&	212	&	VO	&	0.48	&	2.75	&	\textsc{affect}	\\ \hline
halogat	&	\begin{tabular}[c]{@{}l@{}}postpone\\ (iterative)\end{tabular}	&	383	&	VO	&	1.46	&	8.8	&	\textsc{change}	\\ \hline
hall	&	hear	&	201	&	OV	&	-1.52	&	-7.23	&	\textsc{perception}	\\ \hline
hallat	&	\begin{tabular}[c]{@{}l@{}}make heard\end{tabular}	&	568	&	no sig.	&	-0.13	&	-1.01	&	\textsc{other}\\ \hline
hall/hallik	&	hear	&	17557	&	OV	&	-0.56	&	-5.35	&	\textsc{perception}	\\ \hline
hamisít	&	\begin{tabular}[c]{@{}l@{}}make a fake\\ version of\end{tabular}	&	334	&	no sig.	&	-0.07	&	-0.48	&	\begin{tabular}[c]{@{}c@{}}\textsc{creation/}\\\textsc{representation}\end{tabular}	\\ \hline
hanyagol	&	neglect	&	177	&	VO	&	0.4	&	2.17	&	\textsc{affect}	\\ \hline
hatástalanít	&	deactivate	&	268	&	no sig.	&	0.32	&	1.97	&	\textsc{change}	\\ \hline
hátráltat	&	hinder	&	1016	&	VO	&	0.63	&	5.16	&	\textsc{affect}	\\ \hline
helyesel	&	approve	&	635	&	VO	&	1.23	&	8.78	&	\begin{tabular}[c]{@{}c@{}}\textsc{evaluation/}\\\textsc{experience}\end{tabular}	\\ \hline
helytelenít	&	disapprove of	&	198	&	VO	&	1.31	&	6.5	&	\begin{tabular}[c]{@{}c@{}}\textsc{evaluation/}\\\textsc{experience}\end{tabular}	\\ \hline
hívogat	&	call	&	110	&	VO	&	0.56	&	2.51	&	\textsc{affect}	\\ \hline
hiányol	&	miss	&	1024	&	VO	&	0.37	&	3.07	&	\begin{tabular}[c]{@{}c@{}}\textsc{evaluation/}\\\textsc{experience}\end{tabular}	\\ \hline
hibáztat	&	blame	&	561	&	OV	&	-1.31	&	-9.01	&	\begin{tabular}[c]{@{}c@{}}\textsc{evaluation/}\\\textsc{experience}\end{tabular}	\\ \hline
hitelesít	&	\begin{tabular}[c]{@{}l@{}}authenticate,\\ validate\end{tabular}	&	420	&	VO	&	0.49	&	3.38	&	\textsc{change}	\\ \hline
hullat	&	shed	&	106	&	no sig.	&	0.23	&	1.03	&	\textsc{affect}	\\ \hline
húzogat	&	pull (iterative)	&	120	&	VO	&	0.51	&	2.38	&	\textsc{affect}	\\ \hline
idéz	&	cite, remind	&	7603	&	no sig.	&	0.14	&	1.35	&	\textsc{other}\\ \hline
igazgat	&	direct, manage	&	414	&	OV	&	-0.53	&	-3.69	&	\textsc{other}\\ \hline
igazol	&	certify	&	3402	&	no sig.	&	-0.05	&	-0.47	&	\textsc{change}	\\ \hline
igényel	&	demand, need	&	8727	&	OV	&	-1.18	&	-11.15	&	\textsc{ownership}	\\ \hline
ihlet	&	inspire	&	785	&	OV	&	-0.86	&	-6.64	&	\textsc{change}	\\ \hline
illet	&	belong to	&	11303	&	OV	&	-3.27	&	-28.62	&	\textsc{ownership}	\\ \hline
illusztrál	&	illustrate	&	481	&	OV	&	-0.52	&	-3.75	&	\begin{tabular}[c]{@{}c@{}}\textsc{creation/}\\\textsc{representation}\end{tabular}	\\ \hline
imád	&	worship, adore	&	4491	&	VO	&	1.74	&	15.72	&	\begin{tabular}[c]{@{}c@{}}\textsc{evaluation/}\\\textsc{experience}\end{tabular}	\\ \hline
imitál	&	imitate	&	171	&	no sig.	&	0.2	&	1.08	&	\begin{tabular}[c]{@{}c@{}}\textsc{creation/}\\\textsc{representation}\end{tabular}	\\ \hline
indukál	&	induce	&	155	&	OV	&	-1.43	&	-6.27	&	\begin{tabular}[c]{@{}c@{}}\textsc{creation/}\\\textsc{representation}\end{tabular}	\\ \hline
ingat	&	destabilize	&	478	&	VO	&	0.66	&	4.67	&	\textsc{change}	\\ \hline
ingerel	&	irritate, nettle	&	166	&	VO	&	0.47	&	2.46	&	\textsc{change}	\\ \hline
inspirál	&	inspire	&	499	&	OV	&	-0.57	&	-4.12	&	\textsc{change}	\\ \hline
irigyel	&	envy	&	613	&	VO	&	1.16	&	8.32	&	\begin{tabular}[c]{@{}c@{}}\textsc{evaluation/}\\\textsc{experience}\end{tabular}	\\ \hline
irritál	&	irritate	&	273	&	VO	&	0.79	&	4.77	&	\textsc{change}	\\ \hline
ismer	&	know (someone)	&	26934	&	VO	&	0.78	&	7.51	&	\begin{tabular}[c]{@{}c@{}}\textsc{evaluation/}\\\textsc{experience}\end{tabular}	\\ \hline
iszik	&	drink	&	3329	&	VO	&	0.23	&	2.11	&	\textsc{ingestion}	\\ \hline
ízesít	&	spice	&	316	&	no sig.	&	-0.09	&	-0.58	&	\textsc{change}	\\ \hline
ízlel	&	taste	&	142	&	VO	&	1.28	&	5.6	&	\textsc{ingestion}	\\ \hline
javít	&	repair	&	4061	&	VO	&	0.79	&	7.28	&	\textsc{change}	\\ \hline
jelent	&	mean	&	20776	&	OV	&	-1.43	&	-13.67	&	\begin{tabular}[c]{@{}c@{}}\textsc{creation/}\\\textsc{representation}\end{tabular}	\\ \hline
jelez	&	indicate	&	5713	&	OV	&	-0.82	&	-7.68	&	\begin{tabular}[c]{@{}c@{}}\textsc{creation/}\\\textsc{representation}\end{tabular}	\\ \hline
\begin{tabular}[c]{@{}l@{}}jelképez/\\ jelképezik\end{tabular}	&	symbolize	&	1303	&	OV	&	-1.39	&	-11.21	&	\begin{tabular}[c]{@{}c@{}}\textsc{creation/}\\\textsc{representation}\end{tabular}	\\ \hline
jellemez	&	characterize	&	4390	&	OV	&	-0.43	&	-4.01	&	\begin{tabular}[c]{@{}c@{}}\textsc{creation/}\\\textsc{representation}\end{tabular}	\\ \hline
\begin{tabular}[c]{@{}l@{}}jellemez/\\ jellemezik\end{tabular}	&	describe	&	2738	&	no sig.	&	-0.17	&	-1.51	&	\begin{tabular}[c]{@{}c@{}}\textsc{creation/}\\\textsc{representation}\end{tabular}	\\ \hline
kárhoztat	&	condemn	&	180	&	no sig.	&	0.29	&	1.6	&	\begin{tabular}[c]{@{}c@{}}\textsc{evaluation/}\\\textsc{experience}\end{tabular}	\\ \hline
károsít	&	impair	&	1022	&	VO	&	0.77	&	6.29	&	\textsc{change}	\\ \hline
kártalanít	&	indemnify	&	134	&	VO	&	0.85	&	3.98	&	\textsc{change}	\\ \hline
kedvel	&	like	&	4989	&	no sig.	&	-0.16	&	-1.46	&	\begin{tabular}[c]{@{}c@{}}\textsc{evaluation/}\\\textsc{experience}\end{tabular}	\\ \hline
kémlel	&	spy on, observe	&	187	&	no sig.	&	-0.12	&	-0.66	&	\textsc{perception}	\\ \hline
képez/képezik	&	form, constitute	&	6752	&	OV	&	-1.49	&	-13.87	&	\begin{tabular}[c]{@{}c@{}}\textsc{creation/}\\\textsc{representation}\end{tabular}	\\ \hline
képvisel	&	represent	&	8610	&	OV	&	-1.26	&	-11.92	&	\begin{tabular}[c]{@{}c@{}}\textsc{creation/}\\\textsc{representation}\end{tabular}	\\ \hline
keres	&	search for	&	22114	&	no sig.	&	-0.11	&	-1.05	&	\textsc{activity}	\\ \hline
\begin{tabular}[c]{@{}l@{}}keresztez/\\ keresztezik\end{tabular}	&	\begin{tabular}[c]{@{}l@{}}cross (someone's\\ way)\end{tabular}	&	413	&	VO	&	0.62	&	4.29	&	\textsc{affect}	\\ \hline
kerülget	&	\begin{tabular}[c]{@{}l@{}}go around\\ (iterative)\end{tabular}	&	373	&	VO	&	0.56	&	3.73	&	\textsc{covering}	\\ \hline
késleltet	&	delay	&	527	&	VO	&	0.75	&	5.41	&	\textsc{change}	\\ \hline
kevesell	&	deem little	&	441	&	VO	&	1.52	&	9.44	&	\begin{tabular}[c]{@{}c@{}}\textsc{evaluation/}\\\textsc{experience}\end{tabular}	\\ \hline
kifogásol	&	expostulate	&	1790	&	no sig.	&	-0.04	&	-0.36	&	\begin{tabular}[c]{@{}c@{}}\textsc{evaluation/}\\\textsc{experience}\end{tabular}	\\ \hline
kímél	&	spare	&	1134	&	no sig.	&	0.14	&	1.16	&	\textsc{change}	\\ \hline
kockáztat	&	risk	&	667	&	OV	&	-0.56	&	-4.31	&	\textsc{other}\\ \hline
kódol	&	encode	&	178	&	OV	&	-1	&	-5.04	&	\begin{tabular}[c]{@{}c@{}}\textsc{creation/}\\\textsc{representation}\end{tabular}	\\ \hline
kommentál	&	comment on	&	2093	&	VO	&	1.24	&	10.77	&	\begin{tabular}[c]{@{}c@{}}\textsc{evaluation/}\\\textsc{experience}\end{tabular}	\\ \hline
kontrollál	&	control	&	179	&	VO	&	0.64	&	3.38	&	\textsc{affect}	\\ \hline
konzervál	&	conserve	&	196	&	VO	&	0.75	&	4.06	&	\textsc{change}	\\ \hline
koordinál	&	coordinate	&	616	&	no sig.	&	0.03	&	0.2	&	\textsc{affect}	\\ \hline
koptat	&	wear down	&	221	&	no sig.	&	0.17	&	1.02	&	\textsc{change}	\\ \hline
\begin{tabular}[c]{@{}l@{}}korlátoz/\\ korlátozik\end{tabular}	&	constrain, limit	&	3468	&	VO	&	0.7	&	6.46	&	\textsc{affect}	\\ \hline
\begin{tabular}[c]{@{}l@{}}koronáz/\\ koronázik\end{tabular}	&	crown, coronate	&	396	&	OV	&	-1.5	&	-9.06	&	\textsc{other}\\ \hline
korrigál	&	correct	&	429	&	VO	&	0.67	&	4.64	&	\textsc{change}	\\ \hline
korszerűsít	&	modernize	&	351	&	VO	&	0.8	&	5.16	&	\textsc{change}	\\ \hline
kortyol	&	take sips of	&	159	&	VO	&	0.53	&	2.72	&	\textsc{ingestion}	\\ \hline
\begin{tabular}[c]{@{}l@{}}körvonalaz/\\ körvonalazik\end{tabular}	&	outline	&	100	&	no sig.	&	0.04	&	0.18	&	\begin{tabular}[c]{@{}c@{}}\textsc{creation/}\\\textsc{representation}\end{tabular}	\\ \hline
kóstál	&	?	&	129	&	OV	&	-4.85	&	-4.84	&	\textsc{other}\\ \hline
köszön	&	thank	&	21747	&	VO	&	2.07	&	19.76	&	\begin{tabular}[c]{@{}c@{}}\textsc{evaluation/}\\\textsc{experience}\end{tabular}	\\ \hline
követ	&	follow	&	12532	&	OV	&	-0.28	&	-2.65	&	\textsc{covering}	\\ \hline
kritizál	&	criticize	&	1385	&	VO	&	0.4	&	3.46	&	\begin{tabular}[c]{@{}c@{}}\textsc{evaluation/}\\\textsc{experience}\end{tabular}	\\ \hline
kutat	&	research	&	1441	&	OV	&	-0.57	&	-4.86	&	\textsc{activity}	\\ \hline
lapátol	&	shovel	&	120	&	VO	&	0.97	&	4.24	&	\textsc{change}	\\ \hline
lapozgat	&	\begin{tabular}[c]{@{}l@{}}turn the pages of\\ (iterative)\end{tabular}	&	172	&	no sig.	&	-0.19	&	-1.01	&	\textsc{activity}	\\ \hline
lassít	&	slow	&	1072	&	VO	&	1.42	&	11.04	&	\textsc{change}	\\ \hline
latolgat	&	\begin{tabular}[c]{@{}l@{}}ponder upon\\ (a decision)\end{tabular}	&	197	&	no sig.	&	-0.34	&	-1.91	&	\textsc{preference}	\\ \hline
legitimál	&	legitimate	&	141	&	VO	&	1.07	&	4.89	&	\textsc{change}	\\ \hline
les	&	expect, watch for	&	913	&	VO	&	0.56	&	4.53	&	\begin{tabular}[c]{@{}c@{}}\textsc{evaluation/}\\\textsc{experience}\end{tabular}	\\ \hline
likvidál	&	liquidate	&	110	&	VO	&	0.6	&	2.67	&	\textsc{change}	\\ \hline
lóbál	&	swing	&	120	&	VO	&	0.48	&	2.22	&	\textsc{affect}	\\ \hline
lobogtat	&	\begin{tabular}[c]{@{}l@{}}wave some canvas\\ (e.g. a flag)\end{tabular}	&	172	&	OV	&	-0.42	&	-2.28	&	\textsc{affect}	\\ \hline
lógat	&	hang	&	236	&	VO	&	1.03	&	5.73	&	\textsc{affect}	\\ \hline
magasztal	&	extol	&	226	&	no sig.	&	0.27	&	1.58	&	\begin{tabular}[c]{@{}c@{}}\textsc{evaluation/}\\\textsc{experience}\end{tabular}	\\ \hline
manipulál	&	manipulate	&	405	&	VO	&	0.95	&	6.28	&	\textsc{change}	\\ \hline
markol	&	grasp, hold onto	&	198	&	no sig.	&	0.2	&	1.15	&	\textsc{affect}	\\ \hline
másol	&	copy	&	382	&	no sig.	&	-0.02	&	-0.14	&	\begin{tabular}[c]{@{}c@{}}\textsc{creation/}\\\textsc{representation}\end{tabular}	\\ \hline
masszíroz	&	massage	&	124	&	no sig.	&	0.1	&	0.47	&	\textsc{affect}	\\ \hline
mellőz	&	ignore	&	663	&	VO	&	0.5	&	3.85	&	\begin{tabular}[c]{@{}c@{}}\textsc{evaluation/}\\\textsc{experience}\end{tabular}	\\ \hline
méltányol	&	appreciate	&	277	&	VO	&	0.71	&	4.36	&	\begin{tabular}[c]{@{}c@{}}\textsc{evaluation/}\\\textsc{experience}\end{tabular}	\\ \hline
mélyít	&	aggravate	&	246	&	VO	&	1.39	&	7.33	&	\textsc{change}	\\ \hline
menedzsel	&	manage	&	145	&	no sig.	&	0.38	&	1.9	&	\textsc{other}\\ \hline
méreget	&	\begin{tabular}[c]{@{}l@{}}measure\\ (iterative)\end{tabular}	&	156	&	VO	&	1	&	4.81	&	\begin{tabular}[c]{@{}c@{}}\textsc{evaluation/}\\\textsc{experience}\end{tabular}	\\ \hline
mereszt	&	\begin{tabular}[c]{@{}l@{}}open (the eyes) wide\end{tabular}	&	165	&	VO	&	0.72	&	3.69	&	\textsc{change}	\\ \hline
\begin{tabular}[c]{@{}l@{}}mérgez/\\mérgezik\end{tabular}	&	poison	&	218	&	VO	&	1.4	&	7.06	&	\textsc{change}	\\ \hline
mérlegel	&	weigh	&	847	&	VO	&	0.47	&	3.8	&	\begin{tabular}[c]{@{}c@{}}\textsc{evaluation/}\\\textsc{experience}\end{tabular}	\\ \hline
minimalizál	&	minimize	&	171	&	VO	&	0.94	&	4.73	&	\textsc{change}	\\ \hline
moderál	&	moderate	&	121	&	VO	&	1.5	&	5.85	&	\textsc{affect}	\\ \hline
modernizál	&	modernize	&	220	&	VO	&	0.64	&	3.65	&	\textsc{change}	\\ \hline
molesztál	&	molest	&	130	&	VO	&	0.54	&	2.56	&	\textsc{affect}	\\ \hline
mormol	&	murmur	&	100	&	OV	&	-0.9	&	-3.68	&	\begin{tabular}[c]{@{}c@{}}\textsc{creation/}\\\textsc{representation}\end{tabular}	\\ \hline
mos	&	wash	&	818	&	VO	&	0.37	&	2.97	&	\textsc{affect}	\\ \hline
motivál	&	motivate	&	639	&	OV	&	-0.28	&	-2.15	&	\textsc{change}	\\ \hline
mozgat	&	make move	&	1140	&	no sig.	&	0.13	&	1.13	&	\textsc{change}	\\ \hline
mutogat	&	point at (iterative)	&	482	&	OV	&	-0.4	&	-2.85	&	\textsc{covering}	\\ \hline
művel	&	cultivate	&	848	&	OV	&	-0.74	&	-5.86	&	\begin{tabular}[c]{@{}c@{}}\textsc{creation/}\\\textsc{representation}\end{tabular}	\\ \hline
nehezít	&	make more difficult	&	2925	&	no sig.	&	-0.03	&	-0.24	&	\textsc{change}	\\ \hline
nélkülöz	&	lack	&	970	&	VO	&	0.67	&	5.45	&	\begin{tabular}[c]{@{}c@{}}\textsc{evaluation/}\\\textsc{experience}\end{tabular}	\\ \hline
nemz	&	father, sire	&	125	&	OV	&	-0.79	&	-3.62	&	\begin{tabular}[c]{@{}c@{}}\textsc{creation/}\\\textsc{representation}\end{tabular}	\\ \hline
nevel	&	grow, cultivate	&	1961	&	OV	&	-0.34	&	-3.05	&	\begin{tabular}[c]{@{}c@{}}\textsc{creation/}\\\textsc{representation}\end{tabular}	\\ \hline
nevesít	&	name, label	&	161	&	no sig.	&	0.24	&	1.25	&	\textsc{other}\\ \hline
nézeget	&	look at	&	1446	&	VO	&	0.45	&	3.85	&	\begin{tabular}[c]{@{}c@{}}\textsc{evaluation/}\\\textsc{experience}\end{tabular}	\\ \hline
növeszt	&	grow	&	204	&	OV	&	-0.88	&	-4.74	&	\begin{tabular}[c]{@{}c@{}}\textsc{creation/}\\\textsc{representation}\end{tabular}	\\ \hline
nyal	&	lick	&	229	&	no sig.	&	0.33	&	1.93	&	\textsc{affect}	\\ \hline
nyalogat	&	lick	&	308	&	no sig.	&	-0.23	&	-1.52	&	\textsc{affect}	\\ \hline
nyitogat	&	open (iterative)	&	117	&	VO	&	0.47	&	2.18	&	\textsc{affect}	\\ \hline
nyomogat	&	press (iterative)	&	205	&	OV	&	-0.37	&	-2.08	&	\textsc{affect}	\\ \hline
nyomkod	&	nyomkod	&	213	&	VO	&	1.18	&	6.17	&	\textsc{affect}	\\ \hline
nyugtalanít	&	make concerned	&	179	&	OV	&	-0.71	&	-3.75	&	\textsc{change}	\\ \hline
nyújtogat	&	stretch	&	105	&	VO	&	0.49	&	2.15	&	\textsc{change}	\\ \hline
okoz/okozik	&	cause	&	22121	&	OV	&	-0.72	&	-6.91	&	\begin{tabular}[c]{@{}c@{}}\textsc{creation/}\\\textsc{representation}\end{tabular}	\\ \hline
ölt	&	stitch	&	571	&	OV	&	-2.22	&	-12.74	&	\textsc{affect}	\\ \hline
olvas	&	read	&	9886	&	VO	&	0.21	&	2.03	&	\textsc{ingestion}	\\ \hline
olvasgat	&	read	&	336	&	VO	&	0.39	&	2.55	&	\textsc{ingestion}	\\ \hline
öntöz	&	irrigate	&	137	&	no sig.	&	-0.34	&	-1.68	&	\textsc{affect}	\\ \hline
őriz	&	guard	&	3985	&	OV	&	-0.25	&	-2.29	&	\textsc{covering}	\\ \hline
orvosol	&	solve, heal	&	650	&	no sig.	&	0.09	&	0.71	&	\textsc{change}	\\ \hline
összegez	&	\begin{tabular}[c]{@{}l@{}}conclude,\\ summarize\end{tabular}	&	335	&	no sig.	&	0.16	&	1.07	&	\textsc{other}\\ \hline
\begin{tabular}[c]{@{}l@{}}összegez/\\ összegezik\end{tabular}	&	summarize	&	733	&	VO	&	1.1	&	8.24	&	\textsc{change}	\\ \hline
összesít	&	summarize	&	280	&	OV	&	-0.42	&	-2.63	&	\textsc{other}\\ \hline
\begin{tabular}[c]{@{}l@{}}osztályoz/\\ osztályozik\end{tabular}	&	classify	&	324	&	VO	&	1.57	&	8.73	&	\textsc{affect}	\\ \hline
óv	&	guard	&	697	&	VO	&	1.23	&	8.96	&	\textsc{covering}	\\ \hline
övez/övezik	&	surround	&	1144	&	OV	&	-0.65	&	-5.42	&	\textsc{covering}	\\ \hline
pártol	&	support	&	345	&	no sig.	&	0.25	&	1.68	&	\textsc{affect}	\\ \hline
\begin{tabular}[c]{@{}l@{}}példáz/\\ példázik\end{tabular}	&	exemplify	&	378	&	OV	&	-0.41	&	-2.78	&	\begin{tabular}[c]{@{}c@{}}\textsc{creation/}\\\textsc{representation}\end{tabular}	\\ \hline
penget	&	pluck (strings)	&	142	&	OV	&	-0.52	&	-2.57	&	\textsc{affect}	\\ \hline
pénzel	&	support via money	&	198	&	no sig.	&	0.1	&	0.58	&	\textsc{affect}	\\ \hline
pihentet	&	make rest	&	159	&	VO	&	0.55	&	2.86	&	\textsc{affect}	\\ \hline
piszkál	&	pick on	&	256	&	no sig.	&	0.17	&	1.06	&	\textsc{affect}	\\ \hline
pontosít	&	specify	&	558	&	VO	&	0.93	&	6.67	&	\textsc{change}	\\ \hline
\begin{tabular}[l]{@{}l@{}}pontoz/\\pontozik\end{tabular}	&	score	&	102	&	no sig.	&	0.24	&	1.05	&	\begin{tabular}[c]{@{}c@{}}\textsc{evaluation/}\\\textsc{experience}\end{tabular}	\\ \hline
pótol	&	replace	&	1792	&	VO	&	0.45	&	3.96	&	\textsc{change}	\\ \hline
preferál	&	prefer	&	466	&	OV	&	-1.4	&	-9.03	&	\textsc{preference}	\\ \hline
privatizál	&	privatize	&	230	&	no sig.	&	-0.05	&	-0.31	&	\textsc{change}	\\ \hline
próbálgat	&	try on	&	302	&	no sig.	&	-0.15	&	-0.94	&	\textsc{activity}	\\ \hline
propagál	&	propagate ?	&	114	&	no sig.	&	-0.28	&	-1.31	&	\textsc{affect}	\\ \hline
provokál	&	provoke	&	324	&	VO	&	0.96	&	5.93	&	\textsc{affect}	\\ \hline
rág	&	chew	&	571	&	VO	&	0.44	&	3.27	&	\textsc{affect}	\\ \hline
rágcsál	&	nibble	&	126	&	no sig.	&	-0.39	&	-1.85	&	\textsc{affect}	\\ \hline
ráncol	&	crease	&	213	&	no sig.	&	0.29	&	1.7	&	\textsc{change}	\\ \hline
rángat	&	pull around	&	369	&	VO	&	0.59	&	3.91	&	\textsc{affect}	\\ \hline
ratifikál	&	ratify	&	307	&	VO	&	1.31	&	7.58	&	\textsc{change}	\\ \hline
ráz	&	shake	&	1594	&	VO	&	0.3	&	2.63	&	\textsc{affect}	\\ \hline
rehabilitál	&	rehabilitate	&	115	&	VO	&	0.55	&	2.52	&	\textsc{change}	\\ \hline
rejteget	&	hide	&	321	&	no sig.	&	-0.18	&	-1.19	&	\textsc{covering}	\\ \hline
\begin{tabular}[c]{@{}l@{}}rendszerez/\\ rendszerezik\end{tabular}	&	organize	&	143	&	VO	&	0.65	&	3.19	&	\textsc{affect}	\\ \hline
reprezentál	&	represent	&	349	&	OV	&	-0.34	&	-2.28	&	\begin{tabular}[c]{@{}c@{}}\textsc{creation/}\\\textsc{representation}\end{tabular}	\\ \hline
\begin{tabular}[c]{@{}l@{}}részletez/\\ részletezik\end{tabular}	&	detail	&	580	&	VO	&	0.32	&	2.41	&	\textsc{change}	\\ \hline
ritkít	&	\begin{tabular}[c]{@{}l@{}}interspace,\\ make sparser\end{tabular}	&	191	&	OV	&	-1.01	&	-5.23	&	\textsc{change}	\\ \hline
rombol	&	destroy	&	518	&	VO	&	0.6	&	4.39	&	\textsc{change}	\\ \hline
roncsol	&	damage	&	131	&	VO	&	1.09	&	4.82	&	\textsc{change}	\\ \hline
ront	&	spoil	&	2735	&	VO	&	0.78	&	7.04	&	\textsc{change}	\\ \hline
rugdos	&	kick (iterative)	&	164	&	no sig.	&	0.12	&	0.65	&	\textsc{affect}	\\ \hline
rühell	&	hate	&	111	&	VO	&	1.18	&	4.8	&	\begin{tabular}[c]{@{}c@{}}\textsc{evaluation/}\\\textsc{experience}\end{tabular}	\\ \hline
semlegesít	&	neutralize	&	278	&	VO	&	0.58	&	3.56	&	\textsc{change}	\\ \hline
serkent	&	stimulate	&	591	&	VO	&	1.51	&	10.18	&	\textsc{change}	\\ \hline
sért	&	offend	&	4902	&	VO	&	0.5	&	4.7	&	\textsc{change}	\\ \hline
sérteget	&	\begin{tabular}[c]{@{}l@{}}hurt, offend\\ (iterative)\end{tabular}	&	114	&	VO	&	1.43	&	5.53	&	\textsc{affect}	\\ \hline
siettet	&	urge	&	271	&	VO	&	1.26	&	7.03	&	\textsc{affect}	\\ \hline
simogat	&	pet	&	697	&	no sig.	&	0.17	&	1.33	&	\textsc{affect}	\\ \hline
sirat	&	mourn	&	422	&	no sig.	&	-0.14	&	-1.01	&	\begin{tabular}[c]{@{}c@{}}\textsc{evaluation/}\\\textsc{experience}\end{tabular}	\\ \hline
skandál	&	\begin{tabular}[l]{@{}l@{}}loudly, repeatedly\\say\end{tabular}	&	225	&	OV	&	-1.66	&	-7.94	&	\begin{tabular}[c]{@{}c@{}}\textsc{creation/}\\\textsc{representation}\end{tabular}	\\ \hline
sokall	&	deem too much	&	423	&	VO	&	1.17	&	7.62	&	\begin{tabular}[c]{@{}c@{}}\textsc{evaluation/}\\\textsc{experience}\end{tabular}	\\ \hline
sokkol	&	shock	&	444	&	VO	&	0.34	&	2.39	&	\textsc{change}	\\ \hline
stabilizál	&	stabilize	&	503	&	VO	&	1.22	&	8.23	&	\textsc{change}	\\ \hline
stimulál	&	stimulate	&	154	&	VO	&	0.79	&	3.92	&	\textsc{change}	\\ \hline
súlyosbít	&	exasperate	&	844	&	no sig.	&	0.15	&	1.23	&	\textsc{change}	\\ \hline
sürget	&	urge	&	2563	&	OV	&	-0.55	&	-4.98	&	\textsc{affect}	\\ \hline
súrol	&	scour	&	462	&	no sig.	&	-0.2	&	-1.44	&	\textsc{affect}	\\ \hline
\begin{tabular}[c]{@{}l@{}}szabályoz/\\ szabályozik\end{tabular}	&	regulate	&	3630	&	no sig.	&	-0.17	&	-1.62	&	\textsc{change}	\\ \hline
szabotál	&	sabotage	&	128	&	VO	&	1.52	&	6.02	&	\textsc{change}	\\ \hline
szagol	&	smell	&	138	&	no sig.	&	0.06	&	0.29	&	\begin{tabular}[c]{@{}c@{}}\textsc{evaluation/}\\\textsc{experience}\end{tabular}	\\ \hline
számolgat	&	enumerate	&	104	&	no sig.	&	0.35	&	1.56	&	\begin{tabular}[c]{@{}c@{}}\textsc{evaluation/}\\\textsc{experience}\end{tabular}	\\ \hline
szankcionál	&	sanction	&	163	&	no sig.	&	0.01	&	0.07	&	\textsc{other}\\ \hline
szaporít	&	breed	&	311	&	VO	&	0.46	&	2.99	&	\begin{tabular}[c]{@{}c@{}}\textsc{creation/}\\\textsc{representation}\end{tabular}	\\ \hline
szaval	&	recite	&	164	&	OV	&	-0.66	&	-3.38	&	\begin{tabular}[c]{@{}c@{}}\textsc{creation/}\\\textsc{representation}\end{tabular}	\\ \hline
szavatol	&	warrant	&	394	&	VO	&	0.54	&	3.69	&	\begin{tabular}[c]{@{}c@{}}\textsc{evaluation/}\\\textsc{experience}\end{tabular}	\\ \hline
szeg	&	nail, frame	&	281	&	OV	&	-0.52	&	-3.22	&	\textsc{other}\\ \hline
\begin{tabular}[c]{@{}l@{}}szegélyez/\\ szegélyezik\end{tabular}	&	frame	&	252	&	OV	&	-1.14	&	-6.36	&	\textsc{covering}	\\ \hline
szélesít	&	broaden	&	197	&	VO	&	0.78	&	4.21	&	\textsc{change}	\\ \hline
szemléltet	&	show, illustrate	&	490	&	OV	&	-1.08	&	-7.41	&	\begin{tabular}[c]{@{}c@{}}\textsc{creation/}\\\textsc{representation}\end{tabular}	\\ \hline
\begin{tabular}[c]{@{}l@{}}szennyez/\\ szennyezik\end{tabular}	&	contaminate	&	418	&	VO	&	1.27	&	8.08	&	\textsc{change}	\\ \hline
szentesít	&	approve	&	310	&	VO	&	1.07	&	6.46	&	\begin{tabular}[c]{@{}c@{}}\textsc{evaluation/}\\\textsc{experience}\end{tabular}	\\ \hline
szerkeszt	&	edit	&	1902	&	OV	&	-1.01	&	-8.77	&	\textsc{other}\\ \hline
szív	&	suck	&	158	&	VO	&	0.41	&	2.14	&	\textsc{affect}	\\ \hline
szid	&	scold	&	864	&	VO	&	0.47	&	3.79	&	\begin{tabular}[c]{@{}c@{}}\textsc{evaluation/}\\\textsc{experience}\end{tabular}	\\ \hline
\begin{tabular}[c]{@{}l@{}}szidalmaz/\\ szidalmazik\end{tabular}	&	scold	&	141	&	VO	&	0.51	&	2.51	&	\begin{tabular}[c]{@{}c@{}}\textsc{evaluation/}\\\textsc{experience}\end{tabular}	\\ \hline
szigorít	&	tighten, toughen	&	725	&	VO	&	2.02	&	13.06	&	\textsc{change}	\\ \hline
szimbolizál	&	symbolize	&	759	&	OV	&	-1.6	&	-11.29	&	\begin{tabular}[c]{@{}c@{}}\textsc{creation/}\\\textsc{representation}\end{tabular}	\\ \hline
szimulál	&	simulate	&	114	&	no sig.	&	-0.35	&	-1.64	&	\begin{tabular}[c]{@{}c@{}}\textsc{creation/}\\\textsc{representation}\end{tabular}	\\ \hline
színesít	&	color	&	574	&	VO	&	0.36	&	2.7	&	\textsc{change}	\\ \hline
színez/színezik	&	color	&	182	&	no sig.	&	-0.02	&	-0.12	&	\textsc{change}	\\ \hline
szív	&	suck	&	904	&	VO	&	0.52	&	4.21	&	\textsc{affect}	\\ \hline
szolgál	&	serve	&	10865	&	OV	&	-1.51	&	-14.31	&	\textsc{other}\\ \hline
szoptat	&	breast-feed	&	103	&	no sig.	&	0.1	&	0.44	&	\textsc{affect}	\\ \hline
\begin{tabular}[c]{@{}l@{}}szorgalmaz/\\ szorgalmazik\end{tabular}	&	urge, hasten	&	2417	&	OV	&	-0.77	&	-6.93	&	\textsc{affect}	\\ \hline
szorongat	&	squeeze	&	443	&	no sig.	&	-0.15	&	-1.07	&	\textsc{affect}	\\ \hline
szponzorál	&	sponsor	&	137	&	no sig.	&	-0.28	&	-1.39	&	\textsc{affect}	\\ \hline
szűkít	&	make narrower	&	639	&	VO	&	1.14	&	8.24	&	\textsc{change}	\\ \hline
szül	&	give birth to	&	1841	&	OV	&	-0.69	&	-6.03	&	\begin{tabular}[c]{@{}c@{}}\textsc{creation/}\\\textsc{representation}\end{tabular}	\\ \hline
tágít	&	broaden	&	244	&	VO	&	1.03	&	5.8	&	\textsc{change}	\\ \hline
taglal	&	elaborate on	&	402	&	OV	&	-1.08	&	-7.01	&	\textsc{activity}	\\ \hline
takar	&	cover	&	1640	&	OV	&	-1.58	&	-12.96	&	\textsc{covering}	\\ \hline
támogat	&	support, donate	&	24224	&	VO	&	0.56	&	5.39	&	\textsc{affect}	\\ \hline
\begin{tabular}[c]{@{}l@{}}tanulmányoz/\\ tanulmányozik\end{tabular}	&	study	&	2199	&	VO	&	0.23	&	2.04	&	\textsc{ingestion}	\\ \hline
tanúsít	&	be a sign of	&	1003	&	OV	&	-1.56	&	-11.8	&	\begin{tabular}[c]{@{}c@{}}\textsc{creation/}\\\textsc{representation}\end{tabular}	\\ \hline
táplál	&	feed (sustain)	&	1177	&	no sig.	&	-0.08	&	-0.65	&	\textsc{affect}	\\ \hline
tapogat	&	\begin{tabular}[c]{@{}l@{}}feel with the\\ fingers, touch\end{tabular}	&	283	&	no sig.	&	-0.31	&	-1.94	&	\textsc{affect}	\\ \hline
\begin{tabular}[c]{@{}l@{}}tárcsáz/\\ tárcsázik\end{tabular}	&	dial	&	126	&	VO	&	0.66	&	3.07	&	\textsc{affect}	\\ \hline
tarkít	&	fleck	&	260	&	no sig.	&	-0.23	&	-1.43	&	\textsc{change}	\\ \hline
tartalmaz	&	contain	&	10791	&	OV	&	-1.26	&	-11.94	&	\textsc{covering}	\\ \hline
\begin{tabular}[c]{@{}l@{}}tartalmaz/\\ tartalmazik\end{tabular}	&	contain	&	17560	&	OV	&	-0.57	&	-5.48	&	\textsc{covering}	\\ \hline
tehermentesít	&	disencumber	&	156	&	VO	&	0.97	&	4.68	&	\textsc{change}	\\ \hline
teljesít	&	fulfil	&	6881	&	VO	&	0.49	&	4.65	&	\textsc{change}	\\ \hline
terhel	&	burden	&	2586	&	OV	&	-0.66	&	-5.97	&	\textsc{affect}	\\ \hline
termel	&	produce	&	1852	&	OV	&	-1.07	&	-9.25	&	\begin{tabular}[c]{@{}c@{}}\textsc{creation/}\\\textsc{representation}\end{tabular}	\\ \hline
tervezget	&	plan	&	195	&	no sig.	&	-0.26	&	-1.45	&	\begin{tabular}[c]{@{}c@{}}\textsc{creation/}\\\textsc{representation}\end{tabular}	\\ \hline
tetéz	&	intensify, pile on	&	340	&	OV	&	-0.65	&	-4.2	&	\textsc{change}	\\ \hline
\begin{tabular}[c]{@{}l@{}}tisztáz/\\tisztázik\end{tabular}	&	clear up	&	1380	&	VO	&	0.99	&	8.3	&	\textsc{change}	\\ \hline
tisztít	&	clean	&	439	&	VO	&	0.64	&	4.43	&	\textsc{affect}	\\ \hline
tökéletesít	&	improve	&	259	&	VO	&	0.65	&	3.89	&	\textsc{change}	\\ \hline
tologat	&	pull	&	137	&	VO	&	0.46	&	2.27	&	\textsc{affect}	\\ \hline
tolerál	&	tolerate	&	252	&	VO	&	0.92	&	5.29	&	\begin{tabular}[c]{@{}c@{}}\textsc{evaluation/}\\\textsc{experience}\end{tabular}	\\ \hline
tömörít	&	compactify	&	341	&	OV	&	-1.59	&	-8.95	&	\textsc{change}	\\ \hline
tompít	&	make blunter	&	260	&	VO	&	0.69	&	4.12	&	\textsc{change}	\\ \hline
tördel	&	break at	&	160	&	no sig.	&	0.25	&	1.33	&	\textsc{affect}	\\ \hline
törölget	&	wipe (iterative)	&	174	&	no sig.	&	-0.12	&	-0.63	&	\textsc{affect}	\\ \hline
torzít	&	distort	&	350	&	VO	&	1.03	&	6.48	&	\textsc{change}	\\ \hline
\begin{tabular}[c]{@{}l@{}}tükröz/\\tükrözik\end{tabular}	&	mirror	&	4684	&	OV	&	-0.55	&	-5.13	&	\begin{tabular}[c]{@{}c@{}}\textsc{creation/}\\\textsc{representation}\end{tabular}	\\ \hline
túr	&	dig	&	126	&	VO	&	0.88	&	3.97	&	\textsc{affect}	\\ \hline
tűr	&	tolerate	&	1431	&	VO	&	1.34	&	11.01	&	\begin{tabular}[c]{@{}c@{}}\textsc{evaluation/}\\\textsc{experience}\end{tabular}	\\ \hline
üdvözöl	&	greet	&	3267	&	VO	&	1.89	&	16.44	&	\begin{tabular}[c]{@{}c@{}}\textsc{evaluation/}\\\textsc{experience}\end{tabular}	\\ \hline
üldöz	&	chase	&	688	&	no sig.	&	0.17	&	1.32	&	\textsc{affect}	\\ \hline
un	&	be bored by	&	1205	&	VO	&	1.97	&	14.55	&	\begin{tabular}[c]{@{}c@{}}\textsc{evaluation/}\\\textsc{experience}\end{tabular}	\\ \hline
ural	&	dominate	&	2044	&	VO	&	0.43	&	3.79	&	\textsc{affect}	\\ \hline
utál	&	hate	&	2240	&	VO	&	1.67	&	14.16	&	\begin{tabular}[c]{@{}c@{}}\textsc{evaluation/}\\\textsc{experience}\end{tabular}	\\ \hline
utánoz	&	imitate	&	594	&	no sig.	&	-0.11	&	-0.82	&	\begin{tabular}[c]{@{}c@{}}\textsc{creation/}\\\textsc{representation}\end{tabular}	\\ \hline
üzemeltet	&	operate	&	1379	&	OV	&	-1.21	&	-10.02	&	\textsc{activity}	\\ \hline
vakar	&	scratch, scrape	&	206	&	VO	&	0.45	&	2.58	&	\textsc{affect}	\\ \hline
választ	&	choose	&	14679	&	OV	&	-0.89	&	-8.49	&	\textsc{preference}	\\ \hline
váltogat	&	\begin{tabular}[c]{@{}l@{}}switch back\\ and forth\end{tabular}	&	256	&	VO	&	1.23	&	6.77	&	\textsc{change}	\\ \hline
variál	&	\begin{tabular}[c]{@{}l@{}}change back\\ and forth\end{tabular}	&	112	&	OV	&	-0.75	&	-3.29	&	\textsc{change}	\\ \hline
vázol	&	sketch	&	503	&	VO	&	0.95	&	6.66	&	\begin{tabular}[c]{@{}c@{}}\textsc{creation/}\\\textsc{representation}\end{tabular}	\\ \hline
véd	&	protect	&	4816	&	no sig.	&	0.05	&	0.46	&	\textsc{covering}	\\ \hline
vedel	&	drink (a lot of)	&	108	&	OV	&	-1	&	-4.17	&	\textsc{ingestion}	\\ \hline
\begin{tabular}[c]{@{}l@{}}védelmez/\\ védelmezik\end{tabular}	&	protect	&	434	&	no sig.	&	0.17	&	1.18	&	\textsc{covering}	\\ \hline
\begin{tabular}[c]{@{}l@{}}véleményez/\\ véleményezik\end{tabular}	&	opine on	&	442	&	VO	&	0.76	&	5.24	&	\begin{tabular}[c]{@{}c@{}}\textsc{evaluation/}\\\textsc{experience}\end{tabular}	\\ \hline
vereget	&	hit at	&	103	&	no sig.	&	0.21	&	0.96	&	\textsc{affect}	\\ \hline
veszélyeztet	&	endanger	&	6502	&	no sig.	&	-0.03	&	-0.28	&	\textsc{change}	\\ \hline
vét	&	\begin{tabular}[c]{@{}l@{}}make a mistake,\\ offend\end{tabular}	&	226	&	OV	&	-2.57	&	-9.23	&	\begin{tabular}[c]{@{}c@{}}\textsc{creation/}\\\textsc{representation}\end{tabular}	\\ \hline
vezérel	&	guide, motivate	&	771	&	OV	&	-0.32	&	-2.55	&	\textsc{affect}	\\ \hline
villant	&	flash	&	142	&	OV	&	-1.04	&	-4.81	&	\begin{tabular}[c]{@{}c@{}}\textsc{creation/}\\\textsc{representation}\end{tabular}	\\ \hline
visel	&	wear, bear	&	11483	&	OV	&	-0.84	&	-8	&	\textsc{covering}	\\ \hline
\begin{tabular}[c]{@{}l@{}}visszhangoz/\\ visszhangozik/\\ visszhangzik\end{tabular}	&	echo	&	124	&	OV	&	-0.78	&	-3.56	&	\begin{tabular}[c]{@{}c@{}}\textsc{creation/}\\\textsc{representation}\end{tabular}	\\ \hline
viszonoz	&	requite	&	581	&	VO	&	1.5	&	10.09	&	\textsc{other}\\ \hline
vizionál	&	envision	&	129	&	OV	&	-1.95	&	-6.83	&	\begin{tabular}[c]{@{}c@{}}\textsc{creation/}\\\textsc{representation}\end{tabular}	\\ \hline
vizsgál	&	examine	&	10098	&	OV	&	-0.64	&	-6.11	&	\textsc{perception}	\\ \hline
vonogat	&	pull	&	297	&	OV	&	-0.53	&	-3.36	&	\textsc{affect}	\\ \hline
zabál	&	gobble	&	173	&	VO	&	0.56	&	2.96	&	\textsc{ingestion}	\\ \hline
zaklat	&	molest	&	400	&	VO	&	0.56	&	3.86	&	\textsc{affect}	\\ \hline
zárol	&	lock (down)	&	509	&	VO	&	0.6	&	4.36	&	\textsc{change}	\\ \hline
zavar	&	disturb	&	2922	&	no sig.	&	0.16	&	1.51	&	\textsc{change}	\\ \hline
zsákmányol	&	prey, obtain	&	330	&	OV	&	-2.8	&	-10.85	&	\textsc{affect}        	\\ \hline

\end{longtable}

\clearpage

\begin{figure}[t!]
 \centering
   \centering
   \includegraphics[width=.8\linewidth]{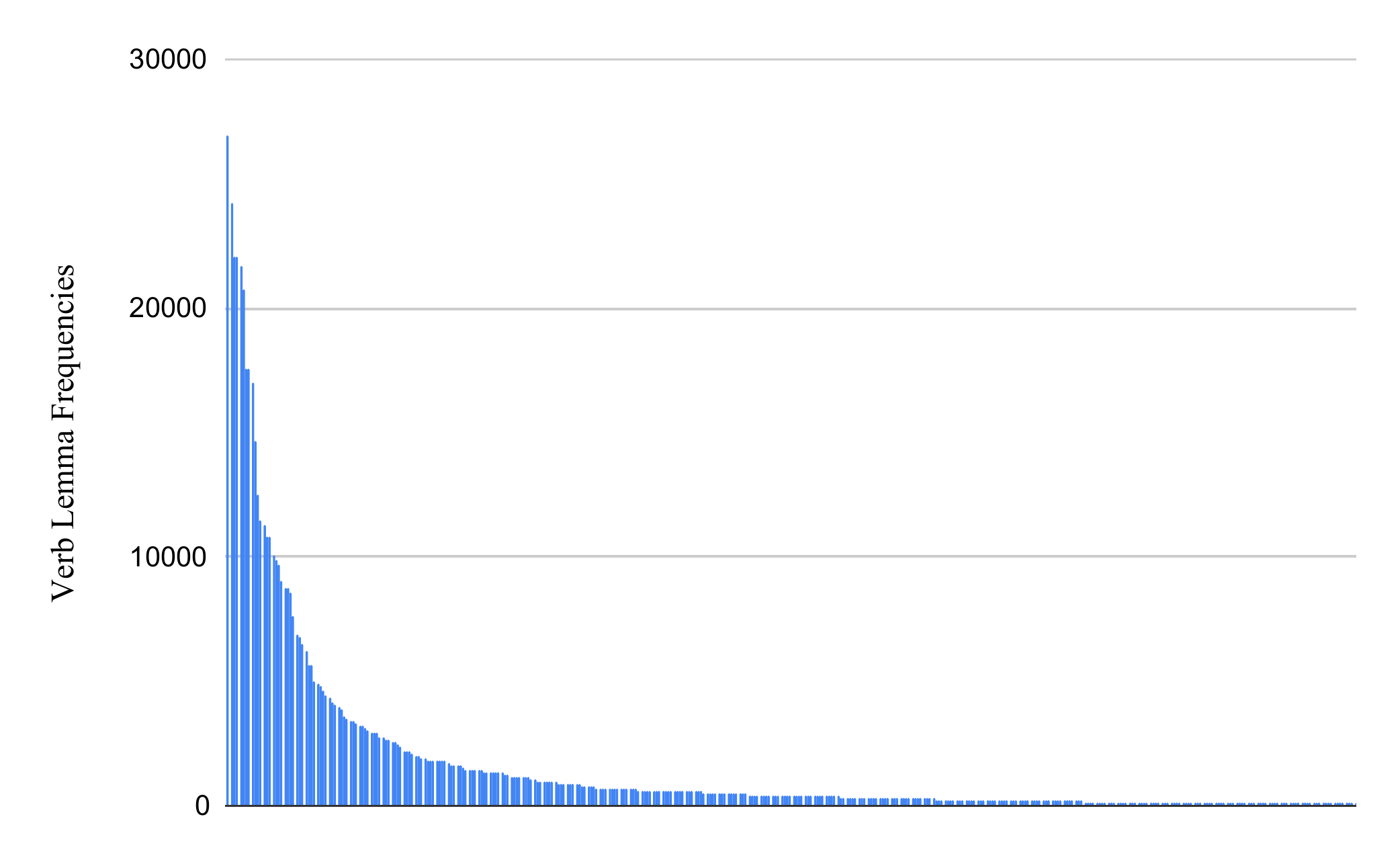}
   \caption{Verb frequencies in our data follow a Zipfian distribution.
   }
   \label{fig:zipf}
\end{figure}

\end{document}